\documentclass[a4paper, 11pt]{article}

\usepackage{arxiv}
\usepackage{microtype}
\usepackage{graphicx}
\usepackage{booktabs}  
\usepackage{makecell}
\usepackage[T1]{fontenc}
\usepackage[utf8]{inputenc}
\usepackage{amsfonts}
\usepackage{amsmath}
\DeclareMathOperator{\E}{\mathbb{E}}

\usepackage{libertine}
\usepackage{bm}
\usepackage{adjustbox}
\usepackage{caption} 
\usepackage{subcaption}
\usepackage{rotating}
\usepackage{hyperref}
\usepackage[capitalise]{cleveref}

\usepackage{pifont}
\usepackage{xcolor}
\newcommand{\cmark}{{\color{green}\ding{51}}}%
\newcommand{\xmark}{{\color{red}\ding{55}}}%

\usepackage[section]{placeins}

\usepackage{geometry}
\usepackage{setspace}
\title{Conditional Wasserstein GAN-based Oversampling\\of Tabular Data for Imbalanced Learning}

\author{
  Justin Engelmann \\
  School of Business and Economics  \\
  Humboldt-Universität zu Berlin \\
  \And
  Stefan Lessmann \\
  School of Business and Economics  \\
  Humboldt-Universität zu Berlin \\ 
}

\begin{document}

\maketitle

\begin{abstract}
Class imbalance is a common problem in supervised learning and impedes the predictive performance of classification models. Popular countermeasures include oversampling the minority class. Standard methods like SMOTE rely on finding nearest neighbours and linear interpolations which are problematic in case of high-dimensional, complex data distributions. Generative Adversarial Networks (GANs) have been proposed as an alternative method for generating artificial minority examples as they can model complex distributions. However, prior research on GAN-based oversampling does not incorporate recent advancements from the literature on generating realistic tabular data with GANs. Previous studies also focus on numerical variables whereas categorical features are common in many business applications of classification methods such as credit scoring. The paper propoes an oversampling method based on a conditional Wasserstein GAN that can effectively model tabular datasets with numerical and categorical variables and pays special attention to the down-stream classification task through an auxiliary classifier loss. We benchmark our method against standard oversampling methods and the imbalanced baseline on seven real-world datasets. Empirical results evidence the competitiveness of GAN-based oversampling. 
\end{abstract}


\section{Introduction}
Data for real-world classification tasks is typically imbalanced. In medical data analysis, for example, a classifier is to detect patients suffering from a disease and those patients will typically represent a minority in the data. Likewise, many marking campaigns reach out to a large number of clients out of which only a few respond. Fraud detection is also known as a "needle in the haystack" problem, and these are only a few examples from the large space of applications of supervised machine learning (ML). Class imbalance is omnipresent. It makes the training of a classifier challenging and can lead to poor predictive performance. Addressing class imbalance is important as ML models are commonly and increasingly used to make many and significant decisions. 

The paper considers credit-related decisions in the financial industry, which heavily relies on ML for decision support. Credit scoring is concerned with estimating the probability that a borrower will not pay back their loan or become otherwise delinquent \cite{lessmann_benchmarking_2015}. Addressing class imbalance is particularly relevant in this setting because data is typically imbalanced as only a small minority of borrowers do not pay back their loans and even small improvements in classification performance can translate to significant financial gains \cite{marques_suitability_2013}. 
 
A common, classification algorithm-agnostic approach to class imbalance is to oversample the data by generating additional synthetic observations of the minority class. However, traditional oversampling approaches rely on nearest neighbours and simple linear interpolations that might fare poorly on high-dimensional, large datasets \cite{chawla_smote_2002}.

Recently, Generative Adversarial Networks (GANs) \cite{goodfellow_generative_2014}, a powerful deep generative technique, have been proposed that can generate complex, high-dimensional data and could, in principle, be used to generate additional minority case examples. While most GAN research focusses on unstructured, continuous data such as images, most real-world classification datasets are tabular, especially in business contexts. Furthermore, tabular data usually contains both numerical and categorical variables. Approaches for using GANs to model tabular data with categorical variables have emerged but this is not yet reflected in the developing, yet sparse literature on GAN-based oversampling.

In this work, we test GAN-based oversampling in the context of imbalanced tabular data for credit scoring. To this end, we develop a GAN architecture that can effectively model tabular data with numerical and categorical variables and pays special attention to the target variable for our downstream classification task, taking into account the state-of-the-art on GAN-based tabular data generation. We then evaluate our proposed GAN-based oversampling method on seven real-world credit scoring datasets and compare against benchmark oversampling methods and against not oversampling using five different classification algorithms. We use data with numerical and categorical variables, whereas the current research on GAN-based oversampling has almost exclusively used purely numerical data. Categorical data is very common in customer facing applications \cite{coussement_dpp}. Credit scorecards routinely embody categorical variables to capture demographic characteristics of an applicant and encoding features of the credit product \cite{baesens_credit_2016}.

Our proposed method outperforms four variants of SMOTE and Random Oversampling on the majority of datasets. However, no oversampling method performs generally better than its alternatives. In fact, not altering the class distribution often delivers competitive results, which indicates that, for the datasets under study, the existence of class skew does not necessarily imply that a data scientist must take actions to address the imbalance. In sum, the paper contributes to the imbalanced learning literature by providing a comprehensive analysis of alternative, recently proposed GAN-based architectures for oversampling and empirical evidence of their performance. The paper also contributes to the credit scoring literature by introducing novel deep learning techniques that might deserve a role in risk analysts toolbox. For example, the problem of low default portfolios is a major obstacle in this field and generative models such as GANs could help to improving the risk management of such portfolios \cite{kennedy_ldp_2013}. 

\section{Background}
Imbalanced learning is concerned with classification on datasets with imbalanced distribution of classes. Class imbalance is a common feature of real-world classification problems as outcomes of interest often occur at different rates. Thus, class imbalance occurs in many different domains \cite{haixiang_learning_2017}. In such a scenario, classifier performance can be adversely affected. Exactly how, and whether, class imbalance adversely affects classification performance depends on the employed classification algorithm \cite{he_learning_2009} \cite{brown_experimental_2012}, the size of the dataset \cite{japkowicz_class_2002}, and on further dataset characteristics that relate to the complexity of the classification task \cite{lopez_insight_2013}. For instance, if classes are easily separable, then class imbalance might not be a problem at all. Conversely, if classes overlap in variable space, or if the minority class occurs in many small clusters, then this makes classification more challenging and in such a scenario, these differences are compounded by class imbalance. Thus, whether class imbalance is a problem in a particular setting cannot be judged by the imbalance ratio alone. 

Class imbalance is a well-studied problem and many approaches have been suggested to address it. Generally, these can be divided into algorithm-level and data-level approaches \cite{yanminsun_classification_2011}. Algorithm-level approaches involve choosing a classification algorithm that is naturally less affected by class imbalance, or choosing an algorithm that is specifically designed for, or adapted to, class imbalance. For example, tree-based models can be adapted by adjusting the probabilistic estimate at each leaf. When information on misclassification costs is available, algorithm-level approaches can be extended to cost-sensitive learning where class imbalance is then implicitly  addressed by minimising the expected misclassification costs. Data-level approaches, on the other hand, address class imbalance by balancing the data through undersampling of the majority class, oversampling the minority class, or a combination thereof. Data-level approaches have the advantage that they can, in principle, be used with any classification algorithm and are relatively straight-forward to apply \cite{yanminsun_classification_2011}. Undersampling is often undesirable as it does not make use of all the available information, especially in settings where the number of minority class samples is low. Oversampling is a popular and promising approach as it makes use of all the available information. 

However, oversampling can lead to overfitting and introduce noise. One of the most popular oversampling approaches is the Synthetic Minority Oversampling TEchnique (SMOTE) \cite{chawla_smote_2002} that generates new minority class samples by linearly interpolating between neighbouring existing ones.
However, in many real-world applications, we have high-dimensional data that includes numerical and categorical variables. Furthermore, for many tabular datasets, the information content of different variables is often heterogeneous with some variables containing vastly more information about the outcome than others. In such a setting, finding nearest neighbours can be problematic as high-dimensional spaces are sparse, and commonly used distance metrics, such as Euclidean distance, are not particularly meaningful. Thus, even the two selected neighbouring minority samples could be conceptually quite different and result in generated synthetic minority cases that might lie in the majority class space. GANs, on the other hand, are able to learn complex, high-dimensional distributions, and thus could potentially perform better in such a setting.

\section{Related literature}
\subsection{GAN-based oversampling}
GANs have been used for data augmentation in the context of unstructured data, such as images \cite{wang_wgan-based_2019}\cite{choi_self-ensembling_2019}. A small literature on using GANs for oversampling structured data has also emerged. However, advancements from the GAN-based tabular data generation literature are not yet reflected, datasets with categorical variables are generally not considered, and comparisons to other oversampling methods are often limited by a small number of datasets, and by the selection of classification algorithms and metrics.

\cite{fiore_using_2019} propose to oversample by training a GAN with the Vanilla GAN loss on only the minority class observations. They compare their method against a SMOTE and no oversampling, and report mixed results. Their comparison is limited, however, as they only use a single classification algorithm, a feed-forward neural network, on a single dataset of credit card fraud which contains only numerical variables, and only use metrics that require a decision threshold. \cite{zheng_generative_2018} propose a complex classification framework which uses a denoising autoencoder in conjunction with a GAN and Gaussian Mixture Models as discriminator and classifier. However, they use weight-of-evidence transformations to convert categorical variables to numerical format and thus do not generate categorical variables with their GAN. Furthermore, while their framework includes a GAN, it is part of a classifier and thus they compare against other classifiers on the original data rather than against other oversampling methods.

\cite{douzas_effective_2017} propose a conditional GAN (cGAN)-based oversampling with the Vanilla GAN loss function and provide a comprehensive evaluation. They compare their method against a number of benchmark oversampling methods on 12 real-world and 10 synthetic datasets. All datasets comprise numerical features whereas categorical features are not present in any of the datasets. \cite{douzas_effective_2017} also create additional versions of the real-world datasets by undersampling the minority class further to yield a total of 71 comparisons. They use five different classification algorithms and three metrics that are suitable for imbalanced settings, and find that that the cGAN performs better than the other methods. This result is encouraging. However, the paper does not report some details of the study design and results in detail, which motivates further research. For one, the discussion of results does not differentiate between real-world and synthetic toy datasets. This is important because the toy datasets create different classes as Gaussian clusters on the vertices of a hypercube. As the cGAN receives Gaussian noise as input, it might simply learn to map this to the vertices of the hypercube effectively and could thus be well-suited for this particular data generation method. Also, the paper  only report mean ranks over all datasets. 
As a consequence, it is hard to appraise on which datasets cGAN performed better and by which margin it outperformed an alternative approach. 

\cite{quintana_towards_2019} adopt two Vanilla GANs and TGAN, short for ``Tabular GAN", a GAN architecture for generating structured data \cite{xu_synthesizing_2018} for oversampling a dataset of subjective thermal comfort which contains categorical and numerical features. However, they only use a single dataset, provide no comparisons with baseline oversampling techniques like SMOTE or Random Oversampling, only use classifiers that are more commonly used in small-n, low dimensional settings (K-Nearest-Neighbours, Support Vector Machine, and Naive Bayes), and report mixed results compared to no oversampling. Still, it is the only study we are aware of where the dataset contains categorical variables and a GAN architecture developed for generation of tabular data was tested.

To summarise, out of the surveyed literature, all but one considered datasets with only numerical variables and no categorical variables and did not pay attention to the quality of the generated data or the available literature on generating realistic tabular data with GANs. The only exception focussed on classifiers for small-n datasets, and did not provide comparisons to other oversampling techniques. Regarding oversampling performance, two studies report mixed success, while one reports promising results.

\subsection{Tabular data generation with GANs}
Although literature on GANs for tabular data is still sparse in comparison to image and language generation, recently a number of approaches have been proposed with the aim of generating realistic synthetic structured data, usually with the aim of creating privacy-preserving, synthetic versions of an existing dataset but without consideration to oversampling.

table-GAN \cite{park_data_2018} is an interesting early work that adapts the widely popular Deep Convolutional GAN (DC-GAN) architecture \cite{radford_unsupervised_2016} from image to structured data instead of designing an architecture specifically for tabular data. To this end, they transform each row from the structured input data into a square two-dimensional matrix which can then be processed by 2d-convolutions. This approach is quite unorthodox, as convolutional operations encode inductive biases into the architecture - namely that relevant patterns can appear anywhere in an image and that neighbouring pixels are more immediately relevant than ones further away. These are appropriate for images but inappropriate for structured data where the order of columns is arbitrary and the information content of columns can be strongly heterogeneous. To deal with categorical variables, table-GAN simply uses label encodings of the categories and rounds the generator output to the nearest integer. Overall, this approach has important  theoretical shortcomings and highlights the need for designing an architecture specifically for tabular data.

medGAN \cite{choi_generating_2018} was developed for a dataset of diagnosis codes associated with patient visits, which leads to a high-dimensional, sparse dataset with only binary columns. It is notable for using the decoder of a pretrained autoencoder to simplify the task of the generator to deal with this unique data structure which appears to work well in this setting. Using a pretrained autoencoder has also been successfully employed for other tabular datasets with more common structures with categorical and numerical attributes, and medGAN-like architectures are often used as a baseline in the literature. \cite{baowaly_synthesizing_2019} report improved results by replacing the Vanilla GAN loss with a Wasserstein GAN Gradient Penalty (WGANGP) loss, which highlights that more recently proposed loss functions that are effective for image generation are also effective in the context of tabular data.

A few architectures have also been proposed for tabular data with both categorical and numerical attributes. TGAN \cite{xu_synthesizing_2018} uses LSTM-Cells with learned attention-weights in the generator to generate columns one-by-one which, in theory, should allow the generator to better model the relationships between columns. However, their follow-up CTGAN \cite{xu_modeling_2019} abandons the LSTM-Cells in favour of ordinary fully-connected layers as the LSTM-Cells are computationally less efficient and did not perform better. They use one-hot encodings and the softmax activation function with added uniform noise to the output for categorical columns and a Gaussian Mixture Model-based approach for numerical columns. The follow-up CTGAN also switches to a WGANGP loss.

\cite{mottini_airline_2018} propose an architecture to generate passenger name record data that is used in the airline industry. Notably, they use crosslayers \cite{wang_deep_2017} in both the generator and the discriminator to explicitly compute feature interactions.
Furthermore, they use categorical embeddings in the discriminator to pass soft one-hot encodings forward to the discriminator which enables them to avoid adding noise like TGAN/CTGAN. They also test Gaussian Mixture Models for numerical columns and find that this reduces the quality of generated data. As loss function they use the multivariate extension of the Cram\'er distance, which has been proposed as an improvement of WGAN \cite{bellemare_cramer_2017}. 

Thus, we find that most of the recent work on generating tabular data uses more recent, advanced GAN loss functions such as WGANGP and can deal with categorical variables, while in the surveyed literature on GAN-based oversampling only the Vanilla GAN loss is used and only one minor work considers categorical variables at all.

Our goal is to use an effective GAN architecture that takes into account the available literature on generating realistic tabular data, especially w.r.t. treatment of categorical variables, to generate high-quality synthetic samples; to compare the results on multiple real-world credit scoring datasets containing both numerical and categorical variables; to benchmark these results against commonly used oversampling techniques with popular classification algorithms; and to present these results in a transparent way.

\section{Methods}

\subsection{Generative Adversarial Networks}
Generative Adversarial Networks (GANs) \cite{goodfellow_generative_2014} are a framework for learning generative models through an adversarial process. Two models are trained together using a training set of real data. One model, the generator $G$, is tasked with generating samples of data that are indistinguishable from the real data, while the other model, the discriminator $D$, tries to discriminate between real examples from the training data and synthetic examples generated by the discriminator. GANs have rapidly gained popularity as a method for modelling complex data distributions and have achieved impressive results, especially in the realm of images \cite{karras_style-based_2019}.

\subsubsection{Vanilla GAN}
If both models are fully differentiable, we can train them with backpropagation. In practice, we use neural networks for both $G$ and $D$, which enables us to approximate complex, high-dimensional distributions. $G$ receives a vector of latent noise drawn from an arbitrary noise distribution $z \sim p_z$ as input and learns to map this to the data space $\mathcal{X}$. $D$ is trained to classify samples as real or fake, while $G$ is trained to minimise $\log (1 - D(G(\bm{z})))$ which corresponds to the following two-player minimax-game
{\small
\begin{equation}
\label{eq:original-gan-objective}
\min_G \max_D \E_{\bm{x} \sim p_{\text{data}}}[\log D(\bm{x})] + \E_{\bm{z} \sim p_{\bm{z}}}[\log (1 - D(G(\bm{z})))] \text{\kern 0.5em .}
\end{equation}%
}%
\cite{goodfellow_generative_2014} also show that given an optimal discriminator the generator's objective is optimised if $p_g$, the generator's distribution over $x$, is equal to the real distribution $p_{\text{data}}$, which corresponds to minimising the Jensen-Shannon Divergence (JSD).

It is common to use a non-saturating version of this loss where the generator maximises $\log (D(G(\bm{z})))$ instead of minimising $\log (1 - D(G(\bm{z})))$, which provides better gradients early in training when the generator is still generating poor samples that the discriminator can reject with high confidence. Still, GANs with the Vanilla GAN loss are difficult to train \cite{goodfellow_nips_2017}. As it is a two-player game, we are not guaranteed to converge to an equilibrium. The losses of $G$ and $D$ do not necessarily correlate with sample quality, which makes it hard to gauge whether the generator is still improving, has converged, or has collapsed. A key problem is mode collapse, where the generator maps all possible values of $z$ to the same output which is the result of solving the maximin-game instead of the minimax-game. This means that the generator finds the single sample that the discriminator classifies as real with the highest confidence.

\subsubsection{Wasserstein GAN and Wasserstein GAN Gradient Penalty}
Wasserstein GAN (WGAN) \cite{arjovsky_wasserstein_2017} aims to address these challenges by optimising the Wasserstein-1 distance instead of the JSD. The Wasserstein-1 distance is also called Earth Mover distance as the Wasserstein-1 distance between two distributions can be interpreted as the ``cost" of the optimal transport plan to move probability mass of one distribution until it matches the other. Thus, even if $p_g$ and $p_{\text{data}}$ have disjoint supports, the Wasserstein-1 distance is a meaningful metric to optimise. If we have two candidate generated distributions $p_g^1$ and $p_g^2$, the Wasserstein-1 distance can tell us which is closer to $p_{\text{data}}$ even neither has any overlap with it. This lies in contrast to the JSD which is infinite for non-overlapping distributions. 

This means that in Vanilla GAN, a powerful discriminator that can distinguish fake and real samples perfectly does not provide useful gradients to the generator and this can lead to the generator collapsing. On the other hand, if the discriminator is less powerful then the generator can successfully fool it without generating realistic samples. Ideally, we would like a powerful discriminator that provides useful gradients for the generator even when the generator output is still poor and this is what WGAN achieves.

Calculating the Wasserstein-1 distance is intractable but it can be approximated by changing the GAN objective to 
{\small
\begin{equation}
\label{eq:wgan-objective}
\min_G \max_D \E_{\bm{x} \sim p_{\text{data}}}[D(\bm{x})] - \E_{\bm{z} \sim p_{\bm{z}}}[D(G(\bm{z}))]
\end{equation}%
}%
as long as $D$ is a $k$-Lipschitz function. This constraint can be satisfied by clipping $D$'s weight to lie in a compact space $[-c, c]$, e.g. $c=0.01$. To implement WGAN with weight-clipping, we need to make two simple modifications to Vanilla GAN: the logs from the loss function are removed and the weights of the discriminator need to be clipped. As the discriminator does not estimate the probability of a given sample being fake, it is often referred to as a ``critic" but we will keep referring to it as the ``discriminator" in this work for consistency. WGAN with weight-clipping generally works better than Vanilla GAN and as the discriminator now approximates a useful distance measure, lower losses during training are generally an indication of good sample quality. 

However, weight-clipping constrains the discriminator's capacity, leads to the weights being pushed to the two extreme values of the permitted range $[-c,c]$ and can lead to exploding or vanishing gradients. WGANGP \cite{gulrajani_improved_2017} aims to remedy these problems by replacing the weight-clipping with a gradient penalty. This makes use of the fact that a differentiable function is $1$-Lipschitz if and only if it has gradients with norm at most $1$ everywhere. Thus, the Lipschitz constraint can be enforced softly by penalising the discriminator with a gradient penalty. The GAN objective thus becomes
{\small 
\begin{equation}
\label{eq:wgangp-objective}
\min_G \max_D \E_{\bm{x} \sim p_{\text{data}}}[D(\bm{x})] - \E_{\bm{z} \sim p_{\bm{z}}}[D(G(\bm{z}))] -  \kern 0.2em \lambda \kern 0.1em \E_{\bm{\hat{x}} \sim p_{\bm{\hat{x}}}}[(\Vert \nabla_{\hat{x}}D(\hat{x}) \Vert_2-1)^2]
\end{equation}%
}%
where $\lambda$ is the penalty coefficient. \cite{gulrajani_improved_2017} recommend using an empirically motivated two-sided penalty that also penalises gradients with norm less than 1. Calculating gradients of $D$ everywhere is intractable and thus the gradients are calculated on linear interpolations $\hat{x} \sim p_{\hat{x}}$ between real and synthetic samples, where $p_{\hat{x}}$ is the sampling distribution of those linear interpolations. The gradient penalty acts as a regularizer and ensures that a discriminator trained to optimality would have a smooth, linear gradient guiding the generator to the data distribution but also constrains the discriminator's capacity.

Although the calculation of the gradient penalty is computationally demanding, WGANGP is easier to train and has become popular for GANs in the domains of images, text and tabular data. While the gradient penalty in WGANGP is well-motivated by theory and needed to satisfy specific theoretical constraints, \cite{fedus_many_2018} show that a gradient penalty can also improve results when added to a Vanilla GAN loss function in a model they call GANGP. This raises some questions about what really drives the performance of WGANGP, but for the purposes of this work, it is sufficient to note that WGANGP works well and that in the context of tabular data generation, the Vanilla GAN loss has fallen out of favour compared to more recent loss functions such as WGANGP.

\subsubsection{Conditional GAN and Auxiliary Classifier GAN}
Another approach to stabilising GAN training is to modify the generator's and discriminator's objectives. Conditional GAN (cGAN) \cite{mirza_conditional_2014} is a simple variation of the GAN objective which conditions the generator on class labels to generate output belonging to a specific class. This is achieved by appending the class label $y$ to both the generator and the discriminator input. Thus, the generator estimates the distribution of $p_{X|y}$ and the discriminator learns to estimate $D(X, y) = P(fake|X, y)$. Conditioning the generator allows us to then sample output belonging to a specific class, while conditioning the discriminator ensures that the generator cannot simply ignore the class label. Conditioning has been shown empirically to make the training process more stable.

Related approaches augment the generator's objective to add a reconstruction loss based on how well an auxiliary model can reconstruct part of the generator input from the generated samples, which forces the generator to produce varied output. Auxiliary Classifier GAN (AC-GAN) \cite{odena_conditional_2017} is an example of this. The generator is conditioned on the class label, but the discriminator is not. Instead, an auxiliary classifier (AC) is introduced that aims to predict the class label of a given sample $AC(X) = P(y|X)$. The cross-entropy loss of the AC is added to the generator loss to encourage the generator to produce samples that recognisably belong to a specific class. In the original formulation, AC and the discriminator share parameters in the hidden layers, which means that the discriminator cannot be conditioned.

\subsection{GANs for tabular data}
By tabular data, also called structured data, we understand a  two-dimensional matrix $X$ where each row represents a sample $x$ and each column represents a variable. There are many differences between tabular data and unstructured data. For instance, the order of columns in tabular data is entirely arbitrary whereas for images and text the order of values of a given sample is of great importance. Furthermore, while in text there is typically one fixed vocabulary of tokens for all positions of a text sample and in images there is a fixed interval of colour intensity values for all pixels of an image, in tabular data there usually is a different vocabulary of possible categories or a different range of possible values for each column. In the present work, we only distinguish numerical and categorical columns as we aim to present a general approach that can be easily applied to different tabular datasets. Thus, we do not distinguish count variables from continuous numerical variables, or between binary, incomparably categorical and ordinal categorical variables. Future work could explore modelling these different column types explicitly. Whereas some work on unstructured data generation does transfer to tabular data, for instance, improved loss functions such as WGANGP have been shown to also be effective for training GANs on tabular data, special considerations are required as it typically consists of heterogeneous data with both numerical and categorical variables, both of which pose specific challenges.

\subsubsection{Modelling categorical variables}\label{cat_treatment}
Categorical variables present a challenge for GANs as both the generator and the discriminator need to be fully differentiable and therefore cannot naturally generate discrete outputs. Thus, it was initially thought that GANs cannot be used to model discrete data \cite{goodfellow_generative_2014} \cite{goodfellow_nips_2017}. However, a number of approaches have since been suggested to overcome this limitation.

While it is possible to generate categorical columns by simply label encoding and treating them like a numerical column during training and then rounding to the nearest integer during sampling \cite{park_data_2018}, it does not perform well in practice and is conceptually questionable, as distance in label space is not related to any sort of conceptual similarity between categories.

A conceptually more sound approach is to one-hot encode each categorical column and apply a softmax activation to the generator output for each categorical column's one-hot columns. However, this presents an issue during the training of the GAN as this produces ``soft" one-hot encodings (e.g. [0.25, 0.50, 0.25]) which the generator can easily distinguish from ``hard" one-hot encodings of the real data samples (e.g. [1, 0, 0]). We could add noise to the hard one-hot vectors of the real data, but unless the softmax class probabilities are close to extremes, we would need to add a lot of noise to make real and synthetic data indistinguishable. Furthermore, we would also expect that adding more noise degrades performance.

When generating synthetic data after training, we can just sample from the soft one-hot encodings to obtain discrete samples. But sampling is not a differentiable operation, and thus cannot be easily used during training. This can be remedied with straight-through estimation \cite{bengio_estimating_2013}, which samples from the soft one-hot encodings and passes hard one-hot encoded samples to the discriminator during the forward-pass, but then calculates the gradients w.r.t. the original soft samples during backpropagation. This works well in practice but leads to biased gradients.

An alternative approach is to use the Gumbel-softmax function \cite{jang_categorical_2017, maddison_concrete_2017} which is a variation of the softmax function where noise from the Gumbel distribution is added to the logits. For a vector of $x$ representing the unnormalised log probabilities for each of the $k$ categories of a categorical variable, the Gumbel-softmax is applied to each element $x_i$ in the following way
{\small \begin{equation}
\text{Gumbel-softmax}(x_{i})=\frac{\exp((x_{i}+g_{i})/\tau)}{\sum_{j=1}^{k}\exp((x_{j}+g_{j})/\tau)}  \text{\kern 2em for \kern 1.5em} i = 1,...,k
\end{equation}%
}%
where $g_1, ..., g_k$ are drawn i.i.d. from $Gumbel(0,1)$ and $\tau$ is a temperature parameter that can be chosen. The resulting output vector is closer to hard one-hot encodings which emulates sampling from the soft one-hot encodings while being fully-differentiable. This also lessens the issues with straight-through estimation as the difference between hard and soft one-hot vectors is smaller, and thus the two can be used well in conjunction. The temperature parameter $\tau$ controls the diversity of the output by making the logits more dissimilar for $\tau < 1$ and more similar for $\tau > 1$.

Another approach is to make it harder to distinguish the generated one-hot-encodings from hard one-hot-encodings by passing them through an embedding layer, instead of aiming to make the two more similar to each other. In the context of language generation, \cite{press_language_2017} propose to use character embeddings and pass the weighted average representation forward instead of the most probable character. This approach can be extended to one-hot encoded categorical columns \cite{mottini_airline_2018}. For each original categorical variable, the corresponding one-hot vector is passed through its own embedding layer, and the embedded representations are input into the discriminator's hidden layers. 

\subsubsection{Modelling numerical variables}\label{num_treatment}

While numerical values are more straight-forward to generate, there are still some issues that need to be considered when generating tabular data. Values in numerical columns are not necessarily real numbers distributed continuously in a fixed interval. We might have integer values, i.e. ``number of items bought" should always be a natural number. More generally, values could occur in specific increments. For instance, we might have a column called ``amount of loyalty points" where all values are a multiple of 10, or a column ``hours rented" where all values are a multiple of 0.25 as rental periods might be intervals of 15 minutes. This might give rise to a similar problem for numerical columns as softmax outputs presented for numerical columns. The discriminator might learn to reject outputs that are not precisely a value that occurs in the original data. But it is difficult for the generator to output exact float values and it would ultimately also not be useful to spend computation time to have the generator learn this.

But even in columns where some values are drawn from a continuous interval, we might still find that there are specific modes that occur with great frequency. Suppose that in a dataset from a telecoms company we have a column called ``total customer spending". If the company offers pay-as-you-go, then the values in that column are continuous. However, some customers might have never incurred any charges and thus there would be many rows where the column is exactly 0. Now suppose the company also offers flat rates, then we will also have many customers where the value is equal to the price of the flat rate. In such a scenario, the generator should learn to generate these specific modes frequently. But generating exact float values is a challenging task for a neural network. Thus, this could lead to the discriminator learning to reject all samples that are close to but not exactly equal to such a frequent mode, as it might be more likely that a value close to a frequent mode, i.e. the flat rate price, stems from an attempt by the generator to output this specific value than that it stems from a real pay-as-you-go customer spending this amount.

To address this issue, we propose to add a small amount of Gaussian noise with zero mean to the numerical columns when training the discriminator. In practice, when we min-max-scale numerical columns to $[0,1]$, we have found that adding noise with a standard deviation of 0.01 works well, although we did not tune this extensively. 
Thus, for a vector $x$ representing the $d$ numerical columns of a sample, we add noise to each element $x_i$ such that $x_i^\text{noisy} = x_i + z_i$, where $z_1, ..., z_d$ are drawn i.i.d. from $\mathcal{N}(0, 0.01)$, and pass the noisy vector on to the discriminator. 
We add the noise to both real and synthetic samples. Adding it to real samples ensures that the discriminator cannot simply reject values that are not precisely equal to a frequent mode of the real data, while adding it to the synthetic samples ensures that the generator is not learning to emulate the noise but instead has an incentive to output values that are identical to a frequent mode in the real data. 

In practice, numerical columns might be correlated with particular realisations of the categorical columns. To allow our generator to model such patterns more easily, we also propose to explicitly condition the numerical output on the categorical output of the generator. We call this approach ``self-conditioning" as the generator output for the numerical output is conditioned on its own output for the categorical columns. This allows the generator to capture relationships between columns more easily, but it should be especially useful when using the Gumbel-softmax activation function since there the categorical output is stochastic and not entirely determined by the generator's hidden state. Without self-conditioning, the generator has to generate the numerical output without knowing which categories were sampled from the categorical output; with self-conditioning, this information is available.

Numerical columns in practice often follow multi-modal distributions. To model this explicitly, using ordinary \cite{xu_synthesizing_2018} and variational \cite{xu_modeling_2019} Gaussian Mixture Models has been proposed. However, this does not address any of the problems outlined above, and it is not strictly necessary to model this explicitly as a sufficiently complex generator can learn to generate multi-modal output natively. \cite{mottini_airline_2018} report that in their setting, using Gaussian Mixture Models degraded performance of their generative model.

\subsection{Our method: cWGAN-based oversampling}\label{our_method}
\subsubsection{GAN objective}\label{our_gan_objective}
To oversample an imbalanced dataset, we first train a GAN to estimate the distribution of our data. Once training is completed, we can oversample the data by using the generator to generate additional samples of the minority class. We use a cGAN structure to estimate the conditional distribution $p_{X|y}$ which allows us to sample the minority class explicitly by conditioning the generator on the minority  class label $x_{\text{new}} = G(z, y=y^\text{minority})$.\footnote{As an alternative to using a cGAN, we considered training a GAN without conditioning to estimate $p_{(X,y)}$ jointly, using the generated class label to identify minority examples. We tested this approach on the UCI adult dataset but found that it performed worse than a cGAN. This is to be expected as $y$ is extremely important in our application but in training a joint GAN it is no more important than any variable of $X$, and due to the class imbalance the generator often fails to generate samples that have a high probability of belonging to the minority class.}

We use the WGANGP loss function due to its theoretical and empirical advantages. To ensure that special attention is paid to the class label during the training process, we further augment our loss function by adding an AC loss to encourage the generator to generate samples that recognisably belong to the given class instead of merely looking plausible given the label. This biases the generator towards generating samples that are especially representative of a specific class. Our approach differs considerably from AC-GAN \cite{odena_conditional_2017}: while in AC-GAN the discriminator and the auxiliary classifier share parameters in the hidden layers, we use two separate networks for both. This allows us to use the AC loss while also conditioning the discriminator, which is not possible with the original AC-GAN setup due to the parameter sharing between D and AC. 

Thus, we have the following two player game
{\small
\begin{equation}
\label{eq:our-objective}
\hspace{-0.3 em}\min_G \max_D 
 \underbrace{\E_{\bm{x} \sim p_{\text{data}}}[D(\bm{x})] - \E_{\bm{z} \sim p_{\bm{z}}}[D(G(\bm{z}))] }_{\text{Wasserstein loss}}
-  
\lambda_{\scriptscriptstyle GP} \underbrace{ \E_{\bm{\hat{x}} \sim p_{\bm{\hat{x}}}}[(\Vert \nabla_{\hat{x}}D(\bm{\hat{x}}) \Vert_2-1)^2] }_{\text{gradient penalty}}
+
 \lambda_{\scriptscriptstyle AC} \underbrace{\E_{\bm{z} \sim p_{\bm{z}}}[BCE(AC(G(\bm{z})))]}_{\text{AC loss}}
\end{equation}%
}%
where $AC$ is our auxiliary classifier network and $BCE$ is the binary cross entropy between the true class label $y \in \{0, 1\}$ and the predicted class probability $y_{pred} \in (0,1)$ defined as $BCE(y, y_{pred}) = - (y \log(y_{pred}) + (1-y) \log(1-y_{pred}))$. 

As the magnitude of the Wasserstein loss fluctuates during training, we scale the AC loss by a scale factor $\lambda_{AC}$ which is continuously updated to ten percent of the absolute value of $D(G(z))$ for the current batch $\lambda_{AC}=0.1 \vert D(G(z)) \vert$ to ensure that minimising the Wasserstein loss is the primary objective of the generator. During backpropagation, the scale factor is treated as a constant. Furthermore, in training we cap the AC loss at $0.3$ so that the generator is not penalised for generating samples where the AC predicts the correct class with at least 74\% confidence to prevent the generator from overfitting on the AC. This means that the generator is not penalised for generating samples that are clearly recognisable as belonging to the conditioned class.

\begin{figure}[!htb] 
\centering
  \includegraphics[width=0.90\linewidth]{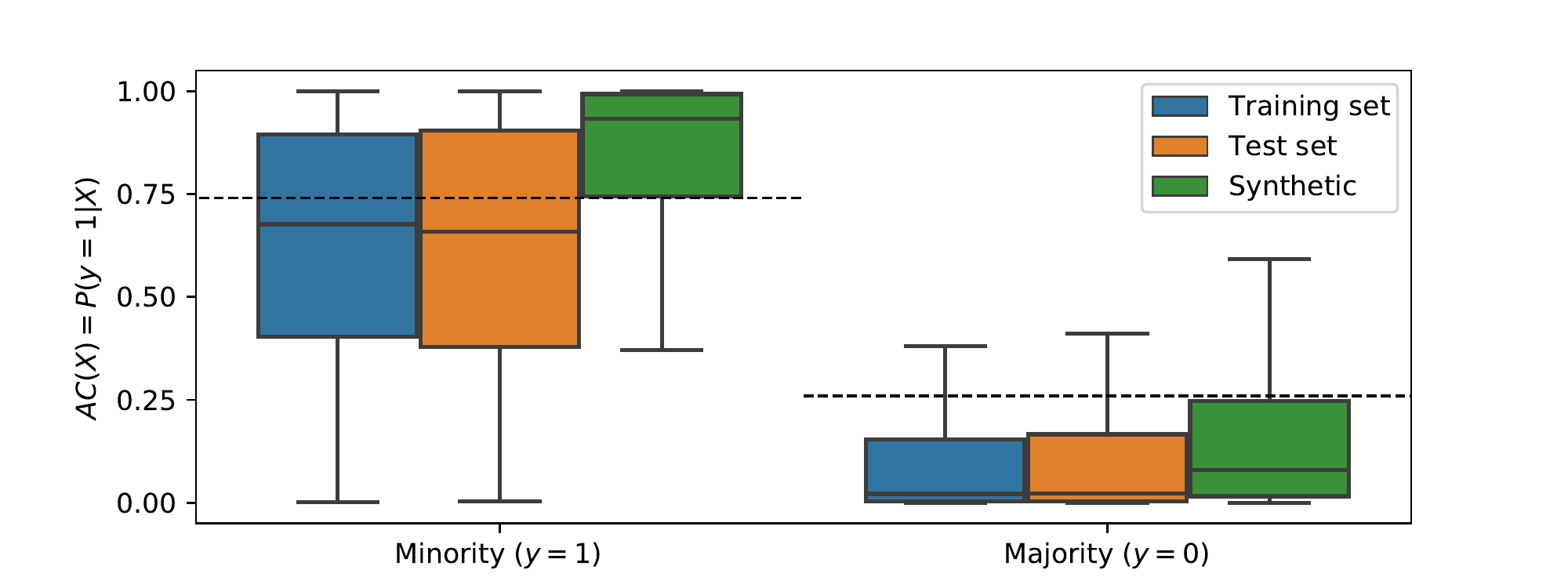}
 \caption[Auxiliary Classifier output predicting minority class membership for samples of the UCI adult dataset]{$AC$ output predicting minority class membership $P(y=1|X)$ for samples of the UCI adult dataset belonging to the minority class (left) and majority class (right). Dashed lines indicate the cut-off values for the AC loss. Outliers are not plotted.}
   \label{fig:AC_preds}
\end{figure}%

To see the effect of this approach on the generated data, we plot the output of the AC for training and test sets of the UCI adult dataset as well as synthetic data generated by a trained generator in  \cref{fig:AC_preds}. This shows that the AC loss leads to the generator generating samples that are clearly recognisable as belonging to the conditioned class. In the case of the harder-to-recognise minority class this leads to synthetic samples that are much easier to recognise than the real minority samples, but the distributions for the majority class show that the generator makes the samples just recognisable enough to avoid the AC loss being applied.

\subsubsection{Modelling tabular data}\label{our_modelling_approaches}
Our GAN architecture is designed to be easily adapted to and effectively model tabular datasets with numerical and categorical columns. We model categorical columns by one-hot encoding them and using the Gumbel-softmax activation function with temperature $\tau = 0.66$. In language modelling, it is common to anneal the temperature during the course of the training to force outputs closer to discrete values. However, we do not anneal the temperature as we do not aim to force the generator output to be discrete as a probability distribution over the categories of a categorical column can be appropriate. We still choose $\tau < 1$ to make outputs more extreme during the initial stages of training to encourage our generator to associate specific values of the noise input $z$ with specific categories. 

For numerical columns, we use min-max-scaling to $[0,1]$ and no activation function for the generator output. We considered using sigmoid and rectified linear units (ReLU) which might seem natural choices in this context and have been used in related work but ultimately performed worse.  A sigmoid function would constrain the output to $(0,1)$ which precludes the generator from generating output equal to the highest and lowest values in the training data, which is undesirable, but also prevent it from generating output that lies outside that range, which is desirable. However, the gradient of the sigmoid function is non-linear and vanishes for extreme values. ReLU precludes the generator from generating values that are smaller than the lowest value in the training data and has linear gradients for any output greater than zero. However, for any values smaller than zero, the derivative of the output is 0 which can lead to the generator collapsing into generating mostly 0 values for numerical columns. Using no activation function means that our generator can generate values outside the range of the training data and has to learn to constrain its own output to mostly this range. Nonetheless, this approach provided the best generative performance in our limited testing on the UCI adult dataset. When sampling from our generator after training is completed, we clip numerical values to $[0,1]$. We also add Gaussian noise to the generator input to address problems outlined in \cref{num_treatment}.

To better model the relationships between variables, we use crosslayers \cite{wang_deep_2017} in both the generator and the discriminator. Crosslayers explicitly calculate feature interactions by calculating
$x_{n+1} =x_0x_n^T w_n + b_n +x_n$ where $x_n$ is the output of the $n$-th crosslayer, $x_0$ is the initial input to the network, and $w_n$ and $b_n$ are the weight and bias of the $n$-th crosslayer, respectively. Stacking $n$ crosslayers allows for efficient computation of $(n+1)$-th degree feature interactions. In the generator, crosslayers increase the variation of the noise, while in the discriminator and the AC they improve their discriminative power. Furthermore, having crosslayers parallel to a deep neural network allows the crosslayers to act similar to a shortcut connection. Shortcut connections are widely used in deep residual networks used for classification \cite{he_deep_2015} and have been used to improve the generator of medGAN \cite{choi_generating_2018}. We also use self-conditioning described in \cref{num_treatment} in the generator to further allow the generator to model interactions between variables and to take into account the random sampling of the categorical output by the Gumbel-softmax function. 

Furthermore, we use embedding layers for dimensionality reduction and better representative power for the input to the discriminator and the AC, as well as for self-conditioning in the generator. An embedding layer is a single weight matrix equivalent to a normal feed-forward layer with the bias frozen at 0. This allows to transform a sparse one-hot encoded vector into a dense vector of real values. In the case of hard one-hot encodings, the weight matrix acts as a look-up table and in the case of soft one-hot encodings generated by the generator, it outputs a weighted average representation which alleviates some of the problems outlined in \cref{cat_treatment}. We initialise the embedding layers at random and train them through backpropagation with the rest of the parameters. Currently, we use separate embedding layers in each network but in the future we would like to explore sharing the embedding weights among the three networks. We set embedding dimensions to $d_{\text{emb}}=min(\left \lceil{k/3}\right \rceil , 20)$ for each categorical column, where $k$ is the number of categories in the column meaning that a sparse one-hot vector of length $k$ is reduced to a dense vector of length $d_{\text{emb}}$.

We sample our noise from a uniform distribution instead of a normal distribution because we use no activation function for numerical outputs. When sampling from a normal distribution, our noise can occasionally take on extreme values which then propagate through the generator and cause extreme values for the numerical output which is exacerbated by also using a crosslayer in the generator. We also use a noise vector of length 30, which is shorter than most values used in the literature that have been focussed on images or inspired by other works focussed on images. Our testing suggests that a short noise vector works better for tabular data. In the future, it might be worth to explore scaling this value with some dataset characteristic, such as the number of clusters identified by a clustering algorithm.

\subsubsection{Network structures}\label{our_network_structures}
\begin{figure}[!htb] %
\centering
\begin{minipage}{.50\textwidth}%
  \centering
  \includegraphics[width=.98\linewidth]{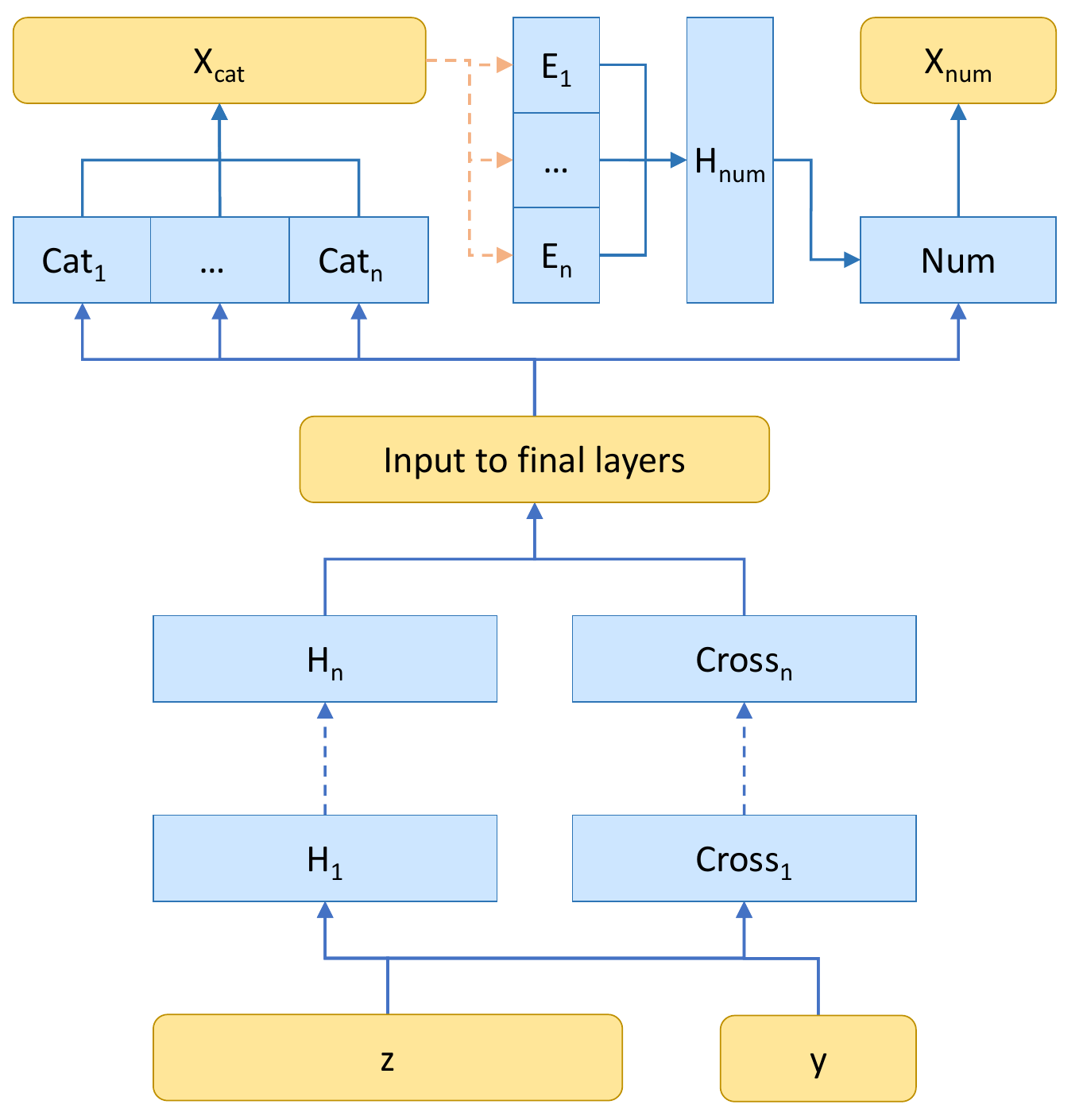}
  \captionof{figure}[Structure of the generator]{Structure of the generator.}
  \label{fig:Gen}
\end{minipage}%
\begin{minipage}{.50\textwidth}%
  \centering
  \includegraphics[width=.98\linewidth]{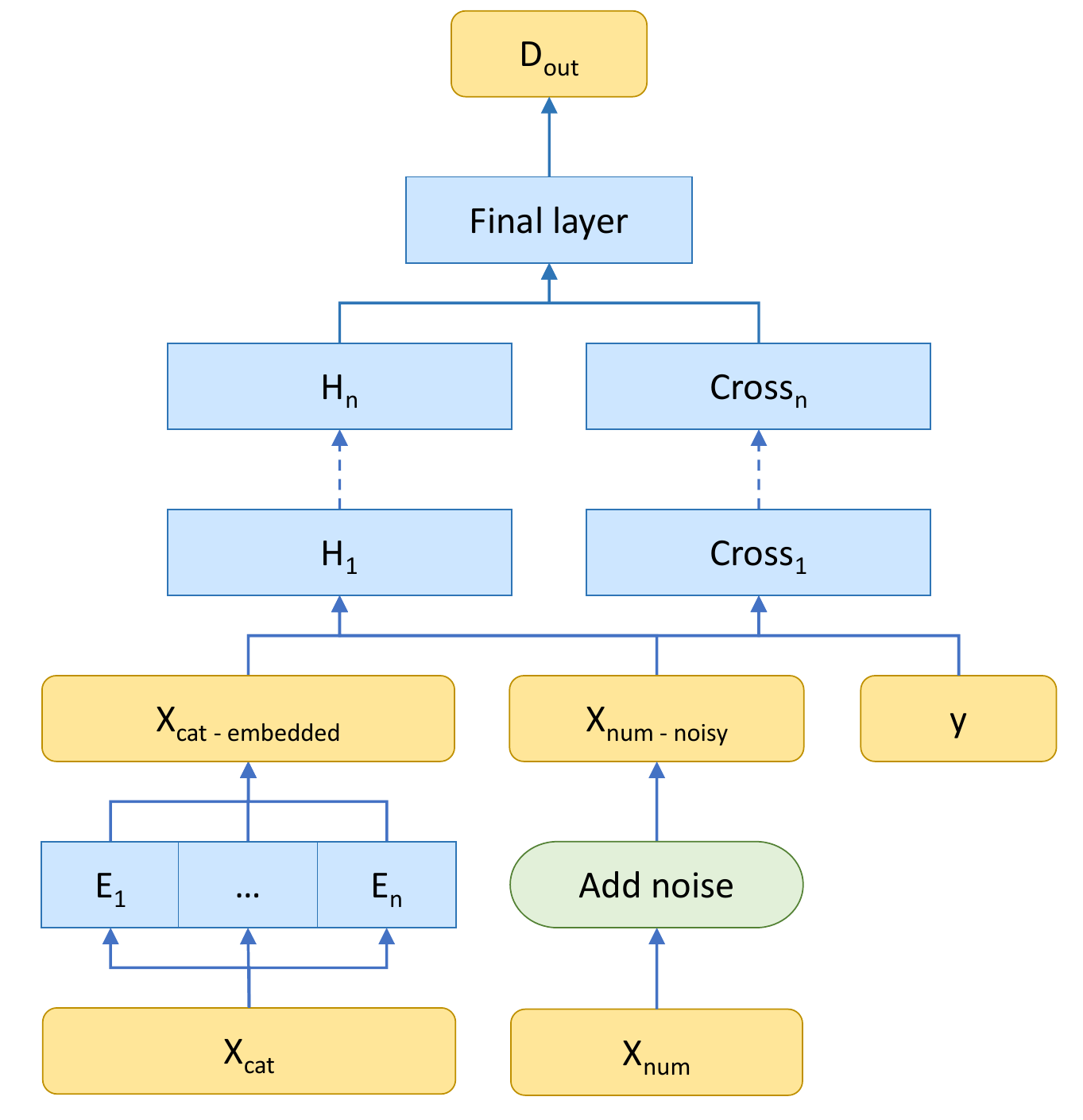}
  \captionof{figure}[Structure of the discriminator]{Structure of the discriminator.}
  \label{fig:Disc}
\end{minipage}%
\end{figure}

The general architectures of our generator and discriminator networks is shown in \cref{fig:Gen,fig:Disc}, respectively.
The discriminator receives a sample of data as input with the associated class label $y$. The categorical columns $X_{cat}$ are embedded by passing the one-hot vector belonging to a categorical column in the original data through the corresponding embedding layer $E_1, ..., E_n$, and Gaussian noise $\mathcal{N}(0, 0.01)$ is added to the numerical columns. The embedded categorical columns and the noisy numerical columns together with $y$ are then passed through both the hidden layers $H_1, ..., H_n$ and the crosslayers $Cross_1, ..., Cross_n$ in parallel. The final layer of the discriminator outputs a scalar, $D_{out}$, and uses no activation function. We use layer normalisation after each hidden layer in the discriminator.

To generate synthetic data, we first take a sample of latent noise $z$ and a class label $y$, which is passed through both the hidden layers $H_1, ..., H_n$ and the crosslayers $Cross_1, ..., Cross_n$ in parallel. The outputs of the final hidden layer and the final crosslayer are then concatenated and form the input to the output layers. We have one output layer for each categorical column $Cat_1, ..., Cat_n$, and one for all numerical columns jointly $Num$. Each categorical output layer $Cat_{i}$ outputs a vector of length $k_{i}$, the number of categories in the $i$-th categorical column. The numerical output layer outputs a vector of length equal to the number of numerical columns. We concatenate the categorical outputs $X_{cat}$ to the input to the numerical output layer to allow the generator to explicitly condition the numerical output $X_{num}$ on $X_{cat}$, but we do not propagate gradients back through this connection. To make this more effective, we first embed each categorical column with an embedding layer and then use a further hidden layer $H_{num}$ to reduce the embeddings of all categorical columns to a single 16-dimensional dense vector.

The AC uses the same general structure as the discriminator with three minor modifications. Firstly, no noise is added to the numerical columns; secondly, the class label is not appended to the input; thirdly, a sigmoid activation function is applied to the output to obtain a class probability. 

\subsubsection{Hyperparameter settings and training schedule}\label{our_gan_hyperparams}
\begin{table}[!ht] %
\label{tab:our_gan_hyperparams}
\centering
\caption[cWGAN hyperparameter settings]{cWGAN hyperparameter settings. Tuned by hand on the UCI adult dataset. Values in curly brackets indicate sets of candidate values to be tuned by grid-search on the training fold.}
\label{tab:tuned-params}
\begin{adjustbox}{max width=\textwidth}
{\small
\begin{tabular}{@{}ll@{}}
\toprule
Hyperparameter                      & Value                                          \\ 
\midrule
Optimizer                           & Adam($\alpha=5 \times 10^{-4}, \beta_1=0, \beta_2=0.9$) \\
Epochs                              & \{300, 500\}                                   \\
Batch size                          & 64      																			\\
Gradient penalty factor $\lambda_{GP}$								      & 15                                             \\
AC loss factor $\lambda_{AC}$								      & 10\% of generator Wasserstein loss on current batch\\
AC loss cut-off		      & 0.3\\
Discriminator updates per generator update          &  3                            \\
Generator layers sizes              & \{(64), (128, 64)\}                            \\
Number of Generator crosslayers     & 1                                              \\
Use extra layer for numerical Generator output & \{yes, no\}                          \\
Discriminator layer sizes           & (128, 64, 32)                                  \\
Number of Discriminator crosslayers & 2                                              \\
AC layer sizes 			                & (64, 64)																				\\
Number of AC crosslayers & 2                                              						\\
Activation function in hidden layers & Leaky ReLU 																	\\
Categorical activation              & Gumbel-softmax($\tau=0.66$)                     \\
Numerical activation                & None                                           \\
Noise distribution $p_z$                                  & $\mathcal{U}^{30}[0,1]$                                  \\ 
Embedding dimension $d_{\text{emb}}$ of a variable with $k$ categories &  $min(\left \lceil{k/3}\right \rceil , 20)$ \\

\bottomrule                                      
\end{tabular}
}
\end{adjustbox}
\end{table}

Our cWGAN model has many hyperparameters that can be chosen. We decided on the parameters in \cref{tab:tuned-params} based on values in the literature on generating tabular data and by tuning on the UCI adult dataset. When ample computation time is available, tuning the fixed parameters via cross-validation will likely improve results. Future work could explore which of these yield the most benefit in the context of oversampling. We train all networks with the Adam optimizer \cite{kingma_adam_2017} with beta values from the original WGANGP paper \cite{gulrajani_improved_2017} and a learning rate $\alpha = 5 \times 10^{-4}$. The AC is pretrained for 30 epochs and then held constant for the rest of the training. 

\subsubsection{Generative performance of our method}
While oversampling performance is our primary objective, we also want to check whether our cWGAN can successfully learn to generate realistic synthetic data. For this purpose, we evaluate the generative performance of the model on the UCI adult dataset that we used for developing our architecture and testing our implementation. No definitive method for measuring the similarity between structured datasets has emerged yet, so we use several different metrics. 

\begin{figure}[!htb] %
\centering
   \includegraphics[width=1\linewidth]{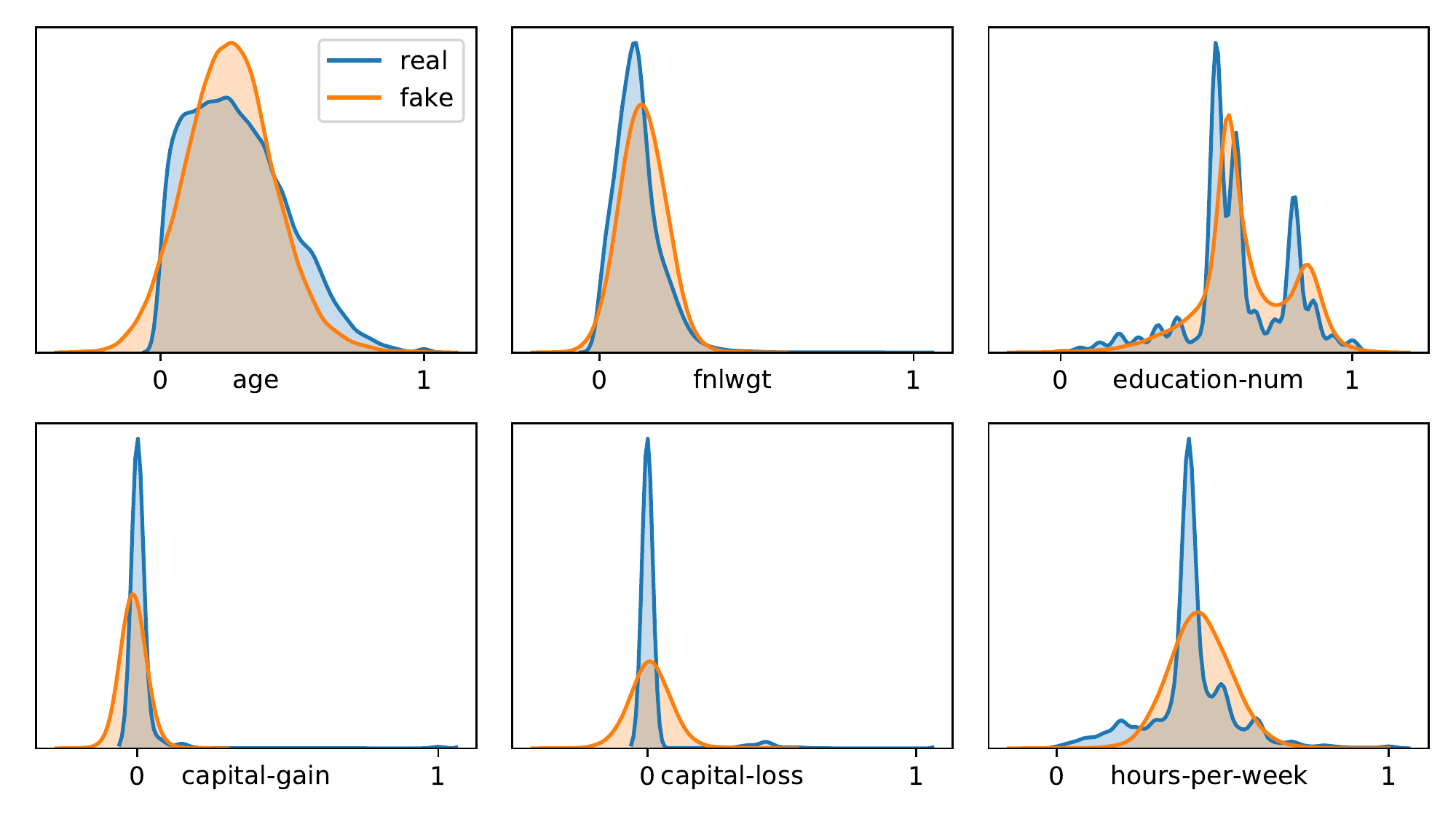}
 \caption[Univariate kernel density plots for the UCI Adult dataset]{Univariate distribution plots comparing the real and generated distributions of the numerical columns of the UCI Adult dataset. We plot Gaussian kernel density estimates with a bandwidth of $0.02$.}
   \label{fig:num_dist_plots_example}
\end{figure}

\begin{figure}[!htb] %
\centering
  \includegraphics[width=1\linewidth]{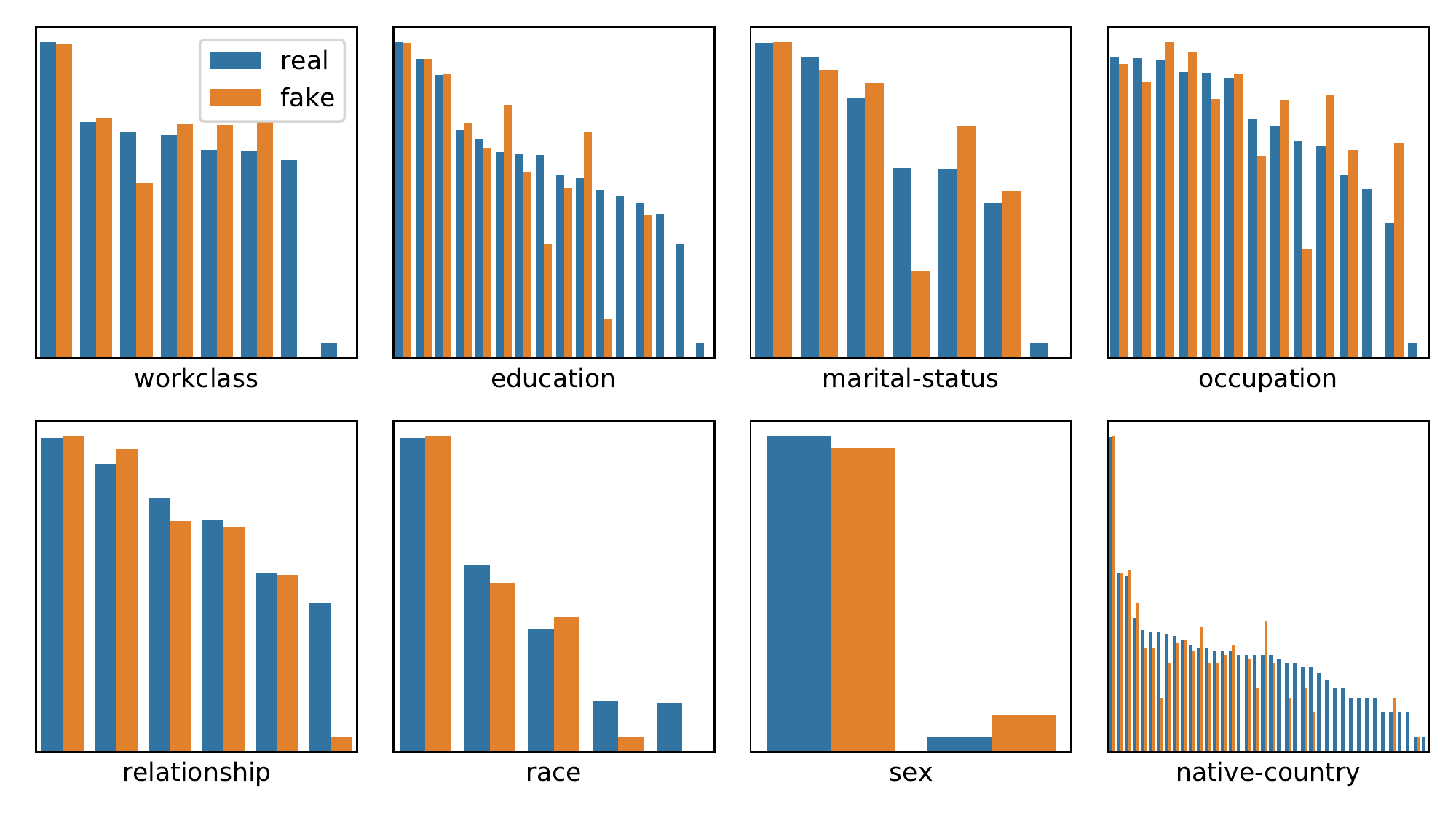}
 \caption[Univariate count plots for the UCI Adult dataset]{Univariate count plots comparing the real and generated distributions of the categorical columns of the UCI Adult dataset. Counts are in log-scale and sorted by frequency in the real data to enable easier comparison for columns with categories that occur infrequently.}
   \label{fig:cat_dist_plots_example}
\end{figure}

First, we visually compare the distributions of each variable individually between synthetic and real data. For numerical variables, we use kernel density estimates with Gaussian kernels and a bandwidth of $0.02$ (\cref{fig:num_dist_plots_example}), and for categorical variables we compare the log-normalized counts for each category (\cref{fig:cat_dist_plots_example}). We find that for numerical columns our cWGAN generally approximates the distribution of individual variables well and manages to capture multiple modes of a distribution but struggles to differentiate between neighbouring modes. For the categorical variables, we find that our cWGAN manages to generate most of the categories present in the real data with approximately the right frequency but fails to generate some of the infrequently occurring categories.

\begin{figure}[!htb] %
\centering
  \centering
  \includegraphics[width=1.01\linewidth]{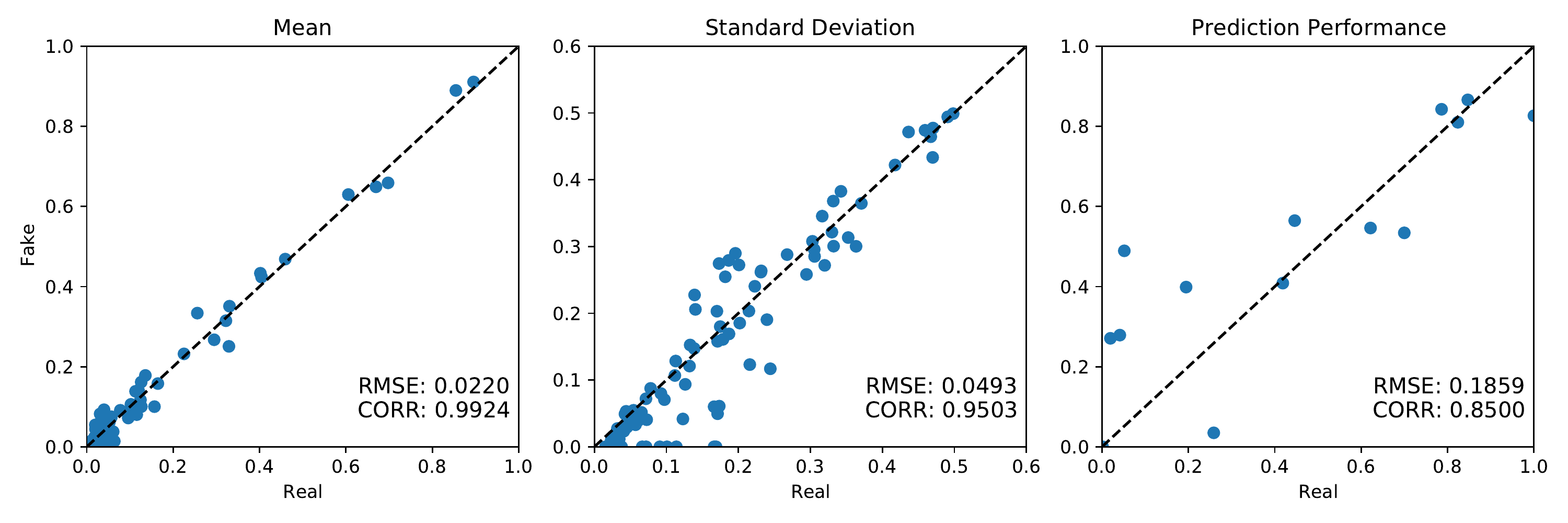}
\caption[Scatterplots of dimension-wise generative performance metrics on the UCI Adult dataset]{Dimension-wise performance metrics on the Adult dataset: dimension-wise means (left), standard deviations (middle), and prediction performance (right).}
  \label{fig:dimwise_plots_example}
\end{figure}

To quantify the generative performance in terms of univariate distributions, we plot a scatterplot of the dimension-wise mean and standard deviation of the synthetic against the real data (\cref{fig:dimwise_plots_example}, left and middle, respectively). \cite{choi_generating_2018} We then quantify the deviation from the identity line with the root mean squared error (RMSE) and also report the Pearson correlation coefficient.  \cite{baowaly_synthesizing_2019} We find that our cWGAN matches both the means and the standard deviations of the original data closely, indicating that the generated data follows the original distribution and does not suffer from mode collapse.

However, these metrics only reflect univariate distributions. To evaluate whether our generator also correctly models the relationships between variables, we also plot dimension-wise prediction performance (\cref{fig:dimwise_plots_example}, right), a metric from the synthetic data generation literature that we extend from numerical variable to both numerical and categorical variables. For both the real and the synthetic datasets, we iterate over all columns. The currently selected column is used as the target variable while the remaining columns are used as covariates to train a supervised prediction model. For categorical columns, we use a random forest model and quantify prediction performance with the weighted $f_1$-score, and for numerical columns, we use a Ridge regression and compute the $R^2$. Models are trained on 90\% of the data and scores are calculated on the remaining 10\%. This metric captures how predictable a column is given the remaining columns, which is a measure of the strength of association between columns. Ideally, those should be identical between real and synthetic data. As with the previous two metrics, we plot values for the real data against those of the synthetic data and summarise it with the RMSE and Pearson correlation coefficient from the identity line. As it is trivial to predict a column of a one-hot vector given all other columns, we use all one-hot columns belonging to the same categorical variable jointly as a target variable. Thus, there are fewer markers on this scatterplot than on the previous two. Modelling the relationships between variables is a challenging objective for our generator and early in training we find that variables are either almost perfectly predictable or unpredictable but this improves during the course of the training.

Overall, we find that our architecture successfully estimates the distribution of a complex, tabular dataset with numerical and categorical attributes and generates synthetic yet realistic data. This is encouraging, especially considering that the AC loss encourages our generator to generate samples that are recognisable as belonging to the conditioned class and thus is biased away from generating modes that occur in the original data but do not clearly belong to a specific class. 

\section{Experimental design}
In order to evaluate the performance of our proposed cWGAN-based oversampling method, we compare our method against four state-of-the-art oversampling methods, Random Oversampling and no oversampling on seven real-world credit scoring datasets.
\subsection{Benchmark methods}
We compare against the baseline performance achieved by training on the original, imbalanced data without addressing class imbalance to test not only which oversampling technique performs best but also whether oversampling improved performance in the first place. In addition to this baseline, we also compare against commonly used oversampling techniques.

Random oversampling is the simplest form of oversampling. Additional minority class samples are generated by drawing with replacement from the original minority class samples. This achieves balanced data but can lead to problematic overfitting as identical samples appear multiple times in the training data and thus a classification model might learn to simply recognise specific minority samples without generalising properly.

SMOTE \cite{chawla_smote_2002} is a popular oversampling approach. New minority class samples $x_{new}$ are generated by choosing a random minority case $x_i$ and then creating a linear combination with a neighbouring minority case $x_j$ randomly selected from the $k$-nearest neighbours on a random point along the line connecting both cases
{\small
\begin{equation}
x_{new} = x_i + \epsilon \times (x_j - x_i)
\end{equation}%
}%
where $\epsilon \sim \mathcal{U}[0,1]$. The number of neighbouring cases to consider $k_\text{neighbours}$ is a hyperparameter.  

SMOTE assumes that all columns are continuous. To deal with categorical variables, \cite{chawla_smote_2002} also propose SMOTE-Nominal Continuous (SMOTENC), a variant of SMOTE for nominal and continuous data. The core idea is to generate the continuous variables of a new minority class sample with SMOTE and then set the nominal features to the most-frequent value in the $k$-nearest neighbours. A slight modification is made to the calculation of the Euclidean distance to penalise differences in nominal variables: first, the median of the standard deviations of all continuous features for the minority class is computed; second, if nominal features differ between a selected minority class example and a potential nearest neighbour, the squared median standard deviation of continuous features is added to the euclidean distance before taking the square root for each nominal feature where the selected sample and potential nearest neighbour differ.

We further consider two variations of SMOTE: Borderline-SMOTE (B-SMOTE) \cite{han_borderline-smote_2005} and ADAptive SYNthetic oversampling (ADASYN) \cite{he_adasyn_2008}. 
B-SMOTE concentrates on borderline minority class samples by first finding the $m$-nearest neighbours of each minority class sample. Samples where the share of majority class samples among the $m$ neighbours is in $[ 0.5, 1)$ are considered to be in danger of being misclassified and used to generate synthetic minority cases with the SMOTE algorithm. Samples below and above this range are considered to be safe and noisy, respectively, and are excluded from being selected for generation of synthetic samples.
ADASYN works similar to SMOTE but selects minority class samples $x_i$ proportionally to the majority class cases in the $k$-nearest neighbours of both classes so more cases are created for minority samples that are neighboured by more majority samples.

In summary, we consider seven oversampling methods for comparison: the six benchmark methods \textit{None}, i.e. not oversampling at all, Random Oversampling, SMOTE, SMOTENC, ADASYN, and B-SMOTE; and cWGAN, our oversampling method based on conditional Wasserstein GAN that we described in section \ref{our_method}.

\subsection{Classification algorithms}
We use five classification algorithms: Random Forest Classifier (RFC), Logistic Regression (Logit), Gradient Boosting Classifier with Decision Trees as base learners (GBC), k-Nearest-Neighbours (KNN), and Decision Tree Classifier (DTC). 
Decision Trees recursively partition the dataset with binary splits selected greedily until each partition only contains samples of a single class. 
Random Forest is an averaging ensemble of Decision Trees, each constructed on a bootstrap subsample of observations and a random subspace of variables to yield uncorrelated trees \cite{breiman_random_2001}. 
Logistic Regression is a linear model that uses a logistic function to model the probabilities of a binary outcome and uses maximum likelihood estimation to fit the model coefficients.
Gradient Boosting builds an additive ensemble by iteratively fitting an additional base model to reduce the error of the current ensemble \cite{hastie_elements_2009}.
k-Nearest-Neighbours classifies a sample by finding the $k$-closest points in the training data and then predicts the class label of the sample via majority vote of the $k$ neighbours. To obtain a class probability rather than a label, we use the fraction of $k$ neighbours that belong to the given class \cite{hastie_elements_2009}. 

Random Forest, Logistic Regression and Gradient Boosting are widely used in academia and industry alike. Logistic Regression performs robustly in a wide range of applications and its coefficients allow for straight-forward interpretability. However, without basis functions and feature engineering, it can only fit a linear decision boundary in variable space and cannot account for interactions. Random Forest and variations of Gradient Boosting with Decision Trees as base learners can learn complex, non-linear decision boundaries while still generalising well to unseen data and often are the best performing classification algorithms on tabular data, especially in business contexts, but offer less interpretability. 

Decision Trees and k-Nearest-Neighbours are used less often in practice as they tend to be outperformed by other techniques \cite{lessmann_benchmarking_2015} but for our evaluation, they provide an interesting measure of oversampling performance as they should be particularly sensitive to the quality of generated samples. Fully-grown, unpruned Decision Trees can learn complex decision boundaries that perfectly fit the training data and associate every point in feature space with a discrete class label but are overfitted and generalise poorly. However, for the purposed of our evaluation, they are interesting: firstly, state-of-the-art classifiers for tabular data such as Random Forest and Gradient Boosting are based on Decision Trees and, secondly, Decision Trees should be especially sensitive to the quality of the generated minority samples. In particular, as the only synthetic samples in the training set are of the minority class, the decision trees could fit on spurious patterns in the generated data, in which case they should perform badly on the test set. Conversely, good performance of an oversampling technique in conjunction with a Decision Tree indicates that the generated synthetic minority samples were realistic. Similarly, k-Nearest-Neighbours is also sensitive to the quality of generated synthetic minority samples but - unlike Decision Trees - 
yields softer and non-orthogonal decision boundaries that are highly dependent on local training examples. Thus, with k-Nearest-Neighbours the oversampling performance depends on whether minority case samples have been generated that lie close to true minority samples in the test set, which is an implicit measure of how well the distribution of the minority class cases has been approximated by the oversampling method. Finally, it has previously been suggested that Decision Trees and k-Nearest-Neighbours suffer from class imbalance in a credit scoring context \cite{brown_experimental_2012} which provides scope for improving performance through oversampling.

\begin{table}[htb]
\caption[Hyperparameter settings for the employed classification algorithms]{Hyperparameter settings for the employed classification algorithms.}
\label{tab:fixed_params_clfs}
\centering
{\small
\begin{tabular}{lll}
\toprule
Algorithm & Hyperparameter & Value \\
\midrule
Random Forest & $n_{\text{trees}}$ & 300 \\
              & $n_{\text{variables per tree}}$ & $\sqrt{n_{\text{variables}}}$ \\
              & Maximum tree depth & Unlimited \\
              & Bootstrap & Yes \\
              &  &  \\
Logistic Regression & Penalty type & L2 / Ridge \\
              & Inverse penalty coefficient $C$ & 10 \\
              &  &  \\
Gradient Boosting & $n_{\text{boosting rounds}}$ & 100 \\
              & Learning rate & 0.1 \\
              & Base learner & Decision Tree \\
              & Maximum tree depth & 3 levels \\
              &  &  \\
k-Nearest-Neighbours & $k_{\text{neighbours}}$ & 5 \\
							& Distance metric & Euclidean \\
              &  &  \\

Decision Tree & Maximum tree depth & Unlimited \\
\bottomrule
\end{tabular}

}
\end{table}%

Due to computational constraints, we use the same set of reasonable default settings for the hyperparameters (\cref{tab:fixed_params_clfs}) for all runs. We regularise the Logistic Regression with an L2-penalty to improve generalisation and numerical stability.

\subsection{Metrics}\label{sec:metrics}
We evaluate the performance of trained classifiers with three metrics: the Area Under the Receiving Operator Curve (AUC-ROC), the Area Under the Precision Recall Curve (AUC-PR), and the Brier score.

The AUC-ROC is the area under the receiving operator characteristic curve which plots the true positive rate against the false positive rate for each binary decision threshold. The true positive rate - also known as recall - is the ratio of positive cases correctly classified as positive to all positive cases. The false positive rate is the ratio of negative cases falsely classified as positive to all negative cases. Intuitively, the AUC-ROC can be interpreted as an approximation of the probability that of a randomly chosen pair of one truly positive and one truly negative sample, the positive sample will have a higher predicted probability of belonging to the positive class. While the AUC-ROC is always in $[0,1]$, an unskilled classifier equivalent to random guessing yields a 45-degree line in ROC-space and thus achieves an expected AUC-ROC of $0.5$.

The AUC-PR summarises the precision-recall curve and provides an estimate of the trade-off between precision and recall at all thresholds. Precision is the ratio of true positives to all samples classified as positive. However, linearly interpolating this area directly provides an estimate of the classification performance that is generally too optimistic \cite{davis_relationship_2006}. Thus, we instead calculate the AUC-PR by taking the mean of precisions achieved at each threshold weighted by the increase in recall over the previous threshold
{\small
\begin{equation}
\text{AUC-PR} = \sum_n (recall_n - recall_{n-1}) \times precision_n
\end{equation}%
}%
where $recall_n$ and $precision_n$ are the recall and precision at the $n$-th threshold.

The Brier score is the mean squared error of predicted probabilities for each outcome and the respective actual outcome
{\small
\begin{equation}
\text{Brier score} = \frac{1}{N}\sum\limits _{i=1}^{N}(pred_i-class_i)^2
\end{equation}%
}%
where $N$ is the number of observations, and $pred_i$ is the predicted probability of $i$ belonging to the positive class and $class_i$ is the true class label. Thus, for the Brier score lower values are better, whereas for the other two metrics higher is better.

We use these metrics for three reasons. First, these metrics are generally well-suited for imbalanced classification problems. Secondly, they require no binary decision-threshold unlike metrics such as accuracy, balanced accuracy, or the $f_1$-score, which is the harmonic mean between precision and recall. As all of the employed classification algorithms, except fully-grown Decision Trees, provide predictions in the form of class probability scores rather than hard class labels, we would need to convert the scores into labels with a binary decision threshold to use such metrics. However, in practice, such a decision threshold would be chosen depending on the error costs involved and might include additional factors, e.g. the loan amount and available collateral in addition to the predicted default risk. A generic threshold like 0.5 would not be suitable for imbalanced classification. This makes it hard to choose a fair threshold for our evaluation that accurately reflects the performance of the trained classification model. Thus, we regard metrics that do not require a threshold as more suitable for our evaluation.\footnote{For example, where the $f_1$-score evaluates the trade-off between precision and recall at a single threshold, the AUC-PR evaluates it at all thresholds. While not all thresholds, especially high ones, are necessarily meaningful for imbalanced classification and credit scoring, it avoids choosing an arbitrary point and provides a fair comparison of performance.}

And finally, each measure provides a slightly different yet relevant metric of classifier performance. The AUC-ROC and the AUC-PR are ranking metrics that capture how well a classifier can separate the two classes. The AUC-ROC weighs both classes equally while the AUC-PR is focussed on the positive class. Both metrics can be suitable when faced with an imbalanced classification problem. However, in practice it is often useful to make decisions based on predicted class probabilities rather than labels or rankings. For that purpose, we need well-calibrated probabilities which is what the Brier score captures. The Brier score is a proper scoring rule, i.e. it is optimised in expectation by reporting the true underlying probability distribution.

\subsection{Datasets}
We compare the oversampling methods on seven real-world retail credit scoring datasets with sample sizes ranging from small to large n obtained from the UCI Machine Learning Repository \cite{dua_uci_2017} \cite{yeh_comparisons_2009}, previous papers from the credit scoring literature\cite{baesens_credit_2016} \cite{thomas_credit_2002}, the Kaggle ``Give Me Some Credit" competition, the 2010 PAKDD data mining challenge, and a commercial peer-to-peer (P2P) lending platform. As we would expect, all of these datasets exhibit moderate to strong class imbalance. Furthermore, with the exception of the Kaggle dataset, they all contain both numerical and categorical columns, highlighting the importance of addressing categorical variables explicitly in our methodology.

\begin{table}[htb]
\caption[Characteristics of the real-world credit scoring datasets]{Characteristics of the real-world credit scoring datasets used for our evaluation.}
\label{tab:datasets}
\begin{adjustbox}{max width=\textwidth}
{\small
\begin{onehalfspacing}
\begin{tabular}{lllllll}
\toprule
        Name &          Source & Samples & Num. columns & Cat. columns & Total categories & Minority class share (\%) \\
\midrule
     Germany &         UCI MLR &    1000 &            7 &           13 &               54 &                        30 \\
 HomeEquity &  Baesens et al. &    5960 &           10 &            2 &               10 &                        20 \\
      Kaggle &          Kaggle &  150000 &           10 &            0 &                0 &                         7 \\
         P2P &    Lending Club &   42506 &           20 &           12 &              117 &                        11 \\
       PAKDD &      PAKDD 2010 &   50000 &            8 &           28 &              386 &                        26 \\
      Taiwan &         UCI MLR &   30000 &           14 &            9 &               77 &                        22 \\
  Thomas &   Thomas et al. &    1225 &           11 &            3 &               18 &                        26 \\
\bottomrule
\end{tabular}

\end{onehalfspacing}
}
\end{adjustbox}
\end{table}
To develop our methodology and test our implementation, we used the UCI Adult dataset with an associated toy binary classification task. This dataset contains 32.000 observations, has numerical and categorical variables, and has a minority class share of 24\%.

\subsection{Procedure}

For each dataset, we conduct six runs per oversampling technique. Firstly, the available data is randomly split into a training set containing 90\% of the data and a test set with the remaining 10\% of the data. We opt for an approach with a large training set instead of crossvalidation as we expect that all oversampling methods have steep learning curves. In particular, having more minority class samples enables SMOTE and SMOTE variants to generate higher quality synthetic samples as it decreases the expected average distance between real minority samples. Having more training data in general can help to reduce the extent of overfitting for Random Oversampling and improve the quality of the generated data for cWGAN-based oversampling.
Secondly, we apply the oversampling technique to the training set to generate new synthetic minority examples, which are added to the training set to balance it. In the present work, we only consider resampling to parity. Thirdly, we use each of the considered classification algorithms to train a model on the balanced training set and obtain predictions for each sample in the test set. Finally, to evaluate the performance, we bootstrap the test set and calculate the metrics 100 times to obtain robust estimates of classification performance.
Each run is conducted with a different random seed, and for a fair comparison we use the same six data partitions across oversampling methods to ensure that test sets are equally difficult for each method, as recommended by  \cite{demsar_statistical_2006}.

For categorical columns, we first replace missing values with the most frequent value and then one-hot encode the column. For numerical columns, we use mean imputation and min-max scale to $[0,1]$. SMOTENC is applied after imputation and scaling but before one-hot encoding. All other methods are applied after preprocessing.

Due to computational constraints, we use the recommended default hyperparameter settings for the SMOTE variants which is $k_{\text{neighbours}}=5$ for all variants and $m_{\text{neighbours}}=10$ for B-SMOTE. For our cWGAN, we primarily use the hyperparameter settings that we tuned on the UCI Adult dataset reported in \cref{tab:tuned-params} but conduct a small cross-validated gridsearch on the training set to allow for some rough adjustment of generator capacity to the complexity to the dataset. In particular, based on the best AUC-ROC score obtained with a Random Forest, we select the number of epochs to train from $\{300, 500\}$, the generator's hidden layer sizes from $\{(64), (128, 64)\}$, and whether to use an extra hidden layer before the numerical output. As our method is both new and complex, we would expect that more extensive fine-tuning for a specific dataset could yield significant performance gains.

\subsection{Implementation}
The code for this project is written in Python 3.8 and \href{https://github.com/ANGELMAN-J/GANbasedOversampling}{our implementation of the proposed cWGAN-based oversampling method is publicly available on Github}. Our cWGAN implementation was written in PyTorch \cite{paszke_pytorch_2019} version 1.5. At the time of writing, all datasets are publicly available and linked to in the source code of the data loading module. 
The Imbalanced-learn package \cite{lemaitre_imbalanced-learn_2017} version 0.6.2 was used for implementations of benchmark oversampling algorithms and the Scikit-learn package \cite{pedregosa_scikit-learn_2011} version 0.22.2 was used for supervised learning algorithms and metrics.

\section{Results}

\subsection{Oversampling benchmark}
\begin{table}[hp]%
\caption[Mean rankings per classifier for each dataset]{Mean rankings per classifier for each dataset. Sorted by best overall rank per dataset. Best rank per dataset and classifier in bold.}
\label{tab:results_rankings_mean}
\centering
\begin{adjustbox}{max width=\textwidth, max totalheight=0.95\textheight}
{\small
\begin{tabular}{llllllll}
\toprule
           & Classifier &  RFC & Logit  & GBC & KNN  &  DTC &       Overall \\
Dataset & Method &               &                   &                  &               &               &               \\
\midrule
Germany & None &           1.7 &      \textbf{1.0} &     \textbf{1.0} &  \textbf{2.3} &           5.0 &  \textbf{2.2} \\
           & SMOTE &  \textbf{1.3} &               3.0 &              5.0 &           5.7 &           2.3 &           3.5 \\
           & B-SMOTE &           3.3 &               4.3 &              6.0 &           3.0 &  \textbf{1.3} &           3.6 \\
           & Random &           5.0 &               4.3 &              3.7 &           6.0 &           2.3 &           4.3 \\
           & ADASYN &           3.7 &               5.3 &              4.0 &           3.7 &           6.0 &           4.5 \\
           & cWGAN &           6.7 &               5.0 &              2.7 &           3.3 &           6.3 &           4.8 \\
           & SMOTENC &           6.3 &               5.3 &              5.7 &           4.0 &           4.7 &           5.2 \\
\midrule
HomeEquity & cWGAN &  \textbf{1.0} &               5.3 &              2.3 &           5.7 &  \textbf{1.3} &  \textbf{3.1} \\
           & SMOTE &           3.0 &               3.7 &              4.3 &           2.7 &           2.0 &  \textbf{3.1} \\
           & None &           4.3 &      \textbf{1.0} &     \textbf{1.0} &           7.0 &           3.7 &           3.4 \\
           & B-SMOTE &           4.3 &               4.7 &              5.0 &           3.3 &           4.0 &           4.3 \\
           & ADASYN &           5.7 &               4.7 &              5.0 &  \textbf{1.3} &           6.0 &           4.5 \\
           & SMOTENC &           2.7 &               5.0 &              7.0 &           3.3 &           5.7 &           4.7 \\
           & Random &           6.7 &               3.7 &              3.3 &           4.7 &           5.3 &           4.7 \\
\midrule
Kaggle & cWGAN &           1.7 &               3.7 &              2.7 &  \textbf{1.3} &  \textbf{2.7} &  \textbf{2.4} \\
           & None &  \textbf{1.0} &               4.3 &     \textbf{1.3} &           2.7 &           3.0 &           2.5 \\
           & B-SMOTE &           3.3 &               2.3 &              4.0 &           2.7 &           3.0 &           3.1 \\
           & Random &           3.0 &      \textbf{2.0} &              3.0 &           5.7 &           4.0 &           3.5 \\
           & SMOTE &           4.7 &               3.7 &              4.3 &           3.3 &           3.0 &           3.8 \\
           & ADASYN &           5.7 &               5.0 &              5.7 &           5.0 &           5.0 &           5.3 \\
\midrule
P2P & None &           2.3 &      \textbf{1.3} &     \textbf{1.7} &           3.7 &           3.7 &  \textbf{2.5} \\
           & cWGAN &           1.7 &               2.3 &              2.7 &           3.3 &           5.3 &           3.1 \\
           & B-SMOTE &           4.3 &               3.7 &              3.7 &  \textbf{2.3} &  \textbf{2.0} &           3.2 \\
           & Random &  \textbf{1.3} &               2.7 &              3.0 &           5.7 &           4.0 &           3.3 \\
           & SMOTE &           4.0 &               4.7 &              5.0 &           4.7 &           3.7 &           4.4 \\
           & SMOTENC &           6.7 &               5.3 &              6.7 &           2.7 &           3.7 &           5.0 \\
           & ADASYN &           5.3 &               6.0 &              5.3 &           5.7 &           5.7 &           5.6 \\
\midrule
PAKDD & None &           2.0 &      \textbf{1.0} &     \textbf{1.3} &  \textbf{1.0} &           5.3 &  \textbf{2.1} \\
           & cWGAN &  \textbf{1.7} &               3.7 &              2.7 &           2.0 &           3.7 &           2.7 \\
           & Random &           3.3 &               2.7 &              3.3 &           4.0 &  \textbf{2.3} &           3.1 \\
           & B-SMOTE &           3.7 &               4.3 &              3.7 &           5.7 &           4.3 &           4.3 \\
           & SMOTE &           4.3 &               5.0 &              4.7 &           5.3 &           3.3 &           4.5 \\
           & SMOTENC &           7.0 &               5.7 &              6.7 &           3.0 &           3.0 &           5.1 \\
           & ADASYN &           5.7 &               4.7 &              5.7 &           7.0 &           6.0 &           5.8 \\
\midrule
Taiwan & None &  \textbf{1.3} &      \textbf{1.3} &     \textbf{2.0} &  \textbf{1.0} &           2.3 &  \textbf{1.6} \\
           & cWGAN &           2.3 &               3.7 &              2.3 &           2.0 &  \textbf{1.7} &           2.4 \\
           & Random &           2.3 &               2.0 &              3.0 &           4.7 &           2.7 &           2.9 \\
           & SMOTE &           4.0 &               4.0 &              3.7 &           3.3 &           3.3 &           3.7 \\
           & ADASYN &           6.0 &               4.7 &              5.3 &           6.7 &           6.0 &           5.7 \\
           & SMOTENC &           6.0 &               6.0 &              5.7 &           4.3 &           7.0 &           5.8 \\
           & B-SMOTE &           5.7 &               6.3 &              6.0 &           6.0 &           5.0 &           5.8 \\
\midrule
Thomas & None &  \textbf{1.7} &      \textbf{2.0} &              3.7 &  \textbf{1.3} &  \textbf{1.0} &  \textbf{1.9} \\
           & Random &           1.7 &               2.0 &              2.3 &           4.3 &           2.0 &           2.5 \\
           & cWGAN &           3.0 &               4.7 &     \textbf{2.0} &           3.7 &           4.3 &           3.5 \\
           & B-SMOTE &           3.7 &               4.7 &              3.7 &           2.0 &           5.7 &           3.9 \\
           & SMOTE &           5.0 &               2.7 &              3.3 &           6.0 &           4.3 &           4.3 \\
           & SMOTENC &           6.7 &               7.0 &              6.0 &           5.3 &           3.7 &           5.7 \\
           & ADASYN &           6.3 &               5.0 &              7.0 &           5.3 &           7.0 &           6.1 \\
\bottomrule
\end{tabular}

}
\end{adjustbox}
\end{table}%

We compare the seven oversampling methods in conjunction with five different classifiers on seven datasets using three metrics of classification performance. This gives us a total of 105 comparisons of the seven oversampling methods. 
To summarise the results, we report the mean ranking aggregated over our three metrics of each technique per classifier for each dataset in \cref{tab:results_rankings_mean}, mean rankings per classifier and per metric in \cref{tab:results_rankings_mean_permetricandclf}, and summarise the relative performance of our method to the benchmarks in \cref{tab:summary_table}.\footnote{For transparency, we report the full, non-aggregated ranks for each metric in \cref{tab:oversampling_ranks} and the raw numerical scores per metric in \cref{tab:oversampling_results_auc_roc,tab:oversampling_results_auc_pr,tab:oversampling_results_brier} in the appendix. Note that SMOTENC is not defined for datasets without categorical columns and thus no SMOTENC results were obtained for the Kaggle dataset which contains only numerical columns.}

Following the recommendation of \cite{demsar_statistical_2006} and \cite{garcia_advanced_2010}, we use the Friedman test with the Iman–Davenport correction to test the null hypothesis that there is no real difference in performance between the tested oversampling methods. The Iman-Davenport correction to Friedman's $\chi_F^2$ is given by $F_F=\frac{(n-1)\chi_F^2}{n(k-1)-\chi_F^2}$, where $k$ is the number of tested methods and $n$ is the number of datasets. $F_F$ is distributed according to the $F$-distribution with $k-1$ and $(k-1)(n-1)$ degrees of freedom. We apply the Friedman test to the ranks of the oversampling methods for each combination of classifier and metric separately. For calculating the means, we assign SMOTENC the rank of SMOTE for the Kaggle dataset. The results are shown in \cref{tab:ftest_table}. 

\begin{table}[!htb]%
\caption[Results for Friedman test with Iman-Davenport correction]{Results for Friedman test with Iman-Davenport correction. Bold indicates $p < 0.05$.}
\centering
\begin{adjustbox}{max width=\textwidth}
{\small
\label{tab:ftest_table}
\begin{tabular}{llll}
\toprule
Classifier & Metric & $F_{(6, 36)}$ &                $p$ \\
\midrule
RandomForest & AUC-ROC &   4.5476 &  \textbf{0.0016} \\
             & AUC-PR &   6.5257 &  \textbf{0.0001} \\
             & Brier\kern 1.2em &   5.0541 &  \textbf{0.0008} \\
Logit \kern 0.8em & AUC-ROC &   7.8345 &  \textbf{0.0000} \\
             & AUC-PR &   6.5838 &  \textbf{0.0001} \\
             & Brier\kern 1.2em &  23.3466 &  \textbf{0.0000} \\
GradientBoosting & AUC-ROC &  31.6552 &  \textbf{0.0000} \\
             & AUC-PR &  12.1798 &  \textbf{0.0000} \\
             & Brier\kern 1.2em &  25.2783 &  \textbf{0.0000} \\
KNN \kern 1em & AUC-ROC &   1.4667 &           0.2173 \\
             & AUC-PR &   1.4445 &           0.2251 \\
             & Brier\kern 1.2em &   4.6924 &  \textbf{0.0013} \\
DecisionTree & AUC-ROC &   2.2508 &           0.0603 \\
             & AUC-PR &   1.7474 &           0.1382 \\
             & Brier\kern 1.2em &  19.4211 &  \textbf{0.0000} \\
\bottomrule
\end{tabular}

}
\end{adjustbox}
\end{table}%
We can reject the null hypothesis of equivalent performance with high confidence for most classifier-metric combinations, with the exception of AUC-ROC and AUC-PR for  both KNN and Decision Tree. This suggests that the Brier score is more sensitive to differences in oversampling performance than the other two metrics, which could be related to the fact that both AUC-ROC and AUC-PR are ranking metrics. KNN and Decision Trees are limited in terms of how many different class probability scores $p$ they can assign: KNN with $k_{\text{neighbours}}=5$ can only assign $p \in \{0.0, 0.2, 0.4, 0.6, 0.8, 1.0\}$, and a fully-grown, unpruned Decision Tree only assigns $p \in \{0, 1\}$, except in rare cases where multiple samples in the training set have the same exact covariates but different class labels. Thus, when evaluated with a ranking metric, they might not be well-suited for differentiating between oversampling methods. 

\begin{table}[htb]%
\centering
\caption[Mean rankings per metric and per classifier, respectively, aggregated over all datasets]{Mean rankings per metric and per classifier, respectively, aggregated over all datasets. Best rank per metric in bold. Sorted by best overall rank.}
\begin{adjustbox}{max width=\textwidth}%
{\small
\begin{onehalfspacing}%
\label{tab:results_rankings_mean_permetricandclf}
\begin{tabular}{llllllllll}
\toprule
{} &   \multicolumn{1}{c}{     Overall }& \multicolumn{5}{c}{Classifier} & \multicolumn{3}{c}{Metric} \\
\cmidrule(lr){2-2} \cmidrule(lr){3-7} \cmidrule(lr){8-10}
{Method} &     Mean Rank &  RFC & Logit & GBC & KNN  &  DTC &       AUC-ROC &        AUC-PR \kern 0.4em  & Brier\\

\midrule
None    &  \textbf{2.3} &  \textbf{2.0} &      \textbf{1.7} &     \textbf{1.7} &  \textbf{2.7} &           3.4 &  \textbf{3.0} &  \textbf{2.3} &     \textbf{1.7} \\
cWGAN   &           3.1 &           2.6 &               4.1 &              2.5 &           3.0 &           3.6 &           3.8 &           3.5 &              2.2 \\
Random  &           3.5 &           3.3 &               2.8 &              3.1 &           5.0 &           3.2 &           3.2 &           3.2 &              4.1 \\
SMOTE   &           3.9 &           3.8 &               3.8 &              4.3 &           4.4 &  \textbf{3.1} &           3.7 &           3.9 &              4.1 \\
B-SMOTE &           4.0 &           4.0 &               4.3 &              4.6 &           3.6 &           3.6 &           3.5 &           4.2 &              4.4 \\
SMOTENC &           5.2 &           5.9 &               5.7 &              6.3 &           3.8 &           4.6 &           5.3 &           5.3 &              5.2 \\
ADASYN  &           5.4 &           5.5 &               5.1 &              5.4 &           5.0 &           6.0 &           5.1 &           5.3 &              5.7 \\
\bottomrule
\end{tabular}

\end{onehalfspacing}%
}
\end{adjustbox}%
\end{table}%

We find that our proposed cWGAN-based oversampling method compares favourably to the state-of-the-art on oversampling. It outperforms all four SMOTE variants on five out of seven datasets and ties with SMOTE on a sixth. Our method also outperforms Random Oversampling on five of the seven datasets. However, not oversampling actually performs robustly and achieves the best mean rank on all but two datasets. On those two, our cWGAN performed the best overall. 

\begin{table}[htb]%
\caption[Summary of the performance of our method relative to benchmark methods]{Summary of the performance of our method relative to benchmark methods.}
\centering
\begin{adjustbox}{max width=\textwidth}
\label{tab:summary_table}
{\small
\begin{onehalfspacing}
\begin{tabular}{llllllll}
\toprule
Dataset & Germany & HomeEquity &  Kaggle &     P2P &   PAKDD &  Taiwan & Thomas \\
\midrule
cWGAN mean rank                        &     4.8 &         3.1 &     2.4 &     3.1 &     2.7 &     2.4 &        3.5 \\
cWGAN rank of mean ranks                    &       6 &         1 * &       1 &       2 &       2 &       2 &          3 \\
cWGAN outperforms ... & & & & & & \\
\multicolumn{1}{r}{... all SMOTE variants?} &  \xmark &    \xmark * &  \cmark &  \cmark &  \cmark &  \cmark &     \cmark \\
\multicolumn{1}{r}{... Random Oversampling?}  &  \xmark &      \cmark &  \cmark &  \cmark &  \cmark &  \cmark &     \xmark \\
\multicolumn{1}{r}{... Not Oversampling?}   &  \xmark &      \cmark &  \cmark &  \xmark &  \xmark &  \xmark &     \xmark \\
cWGAN performs best?   &  \xmark &    \cmark * &  \cmark &  \xmark &  \xmark &  \xmark &     \xmark \\
Not Oversampling performs best?        &  \cmark &      \xmark &  \xmark &  \cmark &  \cmark &  \cmark &     \cmark \\
\bottomrule
\smallskip *  tied with SMOTE
\end{tabular}

\end{onehalfspacing}
}
\end{adjustbox}
\end{table}%

Looking at the mean rankings per classifier and per metric across all datasets, a similar picture presents itself. Not oversampling performs best for each metric and for each classifier, except Decision Trees where SMOTE performed best overall. Our cWGAN method performed consistently well, except when used in conjunction with Logistic Regression.

Among the tested benchmark oversampling methods, Random Oversampling performed best overall. B-SMOTE outperforms SMOTE on at least some individual datasets, but ADASYN is consistently outperformed by SMOTE and ranks last on four of seven datasets. This indicates that the more recent variations of SMOTE do not necessarily outperform their predecessor, mirroring previous findings in the credit scoring literature that indicated that more recent classification algorithms do not necessarily outperform their predecessors \cite{lessmann_benchmarking_2015}.

The strong performance of not oversampling might be explained by the observation that most of the examined datasets are mostly linear, meaning that the two classes are linearly separable, and thus non-linear classifiers do not significantly outperform linear models like Logistic Regression, confirming earlier findings in the credit scoring literature \cite{baesens_benchmarking_2003}. For the following discussion, we consider a dataset to be strongly non-linear if the difference in AUC-ROC obtained on the original, imbalanced data by a Random Forest exceeds that of a Logistic Regression by at least $0.1$. With this definition, only two datasets, Home Equity and Kaggle, are strongly non-linear. In \cref{tab:summary_table_non}, we contrast dataset linearity with performance of our cWGAN and no oversampling.

\begin{table}[!htb]%
\caption[Linearity of datasets and performance of our method relative to no oversampling]{Linearity of datasets and performance of our method relative to no oversampling.}
\centering
\resizebox{\linewidth}{!}{%
\label{tab:summary_table_non}
{\small
\begin{onehalfspacing}
\begin{tabular}{llllllll}
\toprule
Dataset & HomeEquity &  Kaggle & Germany &   PAKDD &  Taiwan &     P2P & Thomas \\
\midrule
AUC-ROC RFC w/o Oversampling &      0.9733 &  0.8415 &  0.7605 &  0.6203 &  0.7702 &  0.7539 &     0.6265 \\
AUC-ROC Logit w/o Oversampling        &      0.7738 &  0.6981 &  0.7455 &  0.6181 &  0.7685 &  0.7630 &     0.6528 \\
Difference                            &      0.1995 &  0.1434 &  0.0150 &  0.0022 &  0.0017 & -0.0091 &    -0.0263 \\
Strongly non-linear?                  &      \cmark &  \cmark &  \xmark &  \xmark &  \xmark &  \xmark &     \xmark \\
\midrule
Not Oversampling performs best?       &      \xmark &  \xmark &  \cmark &  \cmark &  \cmark &  \cmark &     \cmark \\
cWGAN outperforms Not Oversampling?   &      \cmark &  \cmark &  \xmark &  \xmark &  \xmark &  \xmark &     \xmark \\
cWGAN performs best?                  &    \cmark * &  \cmark &  \xmark &  \xmark &  \xmark &  \xmark &     \xmark \\
\bottomrule
\smallskip *  tied with SMOTE
\end{tabular}

\end{onehalfspacing}
}%
}%
\end{table}%

On the two non-linear datasets, our cWGAN was the best performing technique, while on the remaining datasets not oversampling at all performed best. This could mean that when a dataset is mostly linear, class imbalance is not a considerable problem, as simpler, linear decision boundaries are appropriate, and thus oversampling is less effective. However, when the dataset is strongly non-linear, more complex, non-linear decision boundaries are optimal, which are hard to learn with imbalanced classes. In such a scenario, our cWGAN can successfully generate realistic synthetic minority class samples even when the distribution is complex. This is also supported by the fact that cWGAN performed well in conjunction with Random Forest and Gradient Boosting but badly when paired with Logistic Regression (\cref{tab:results_rankings_mean_permetricandclf}). When the dataset is mostly linear, using a linear model like Logistic Regression is often preferable as it is more easily interpretable. Based on our results, we would expect that cWGAN-based oversampling will not improve classification performance in such a scenario. In fact, it is likely that oversampling would be unnecessary in the first place. However, when the dataset is strongly non-linear, then using a tree-based model like Random Forest or Gradient Boosting is advisable and oversampling could improve classification performance. In such a case, cWGAN-based oversampling could be the best oversampling method as it can effectively estimate even complex data distributions.

\FloatBarrier
\subsection{Ablation study}%
Our proposed cWGAN-based oversampling is a complex method with many elements that are not present in the existing GAN-based oversampling literature, such as the WGANGP loss, the AC loss, and special considerations for treating categorical columns. While our method has additional complexities, we consider these three to be the most important elements. To evaluate whether they were indeed critical to the performance of our method, we provide ablations in \cref{tab:ablation}. We test running our method without the WGANGP loss (using the Vanilla GAN loss instead), without the AC loss, without both, and without special considerations for categorical columns (treating each one-hot encoded column like a numerical column instead). Due to computational constraints, we only provide ablations for the two datasets where cWGAN-based oversampling performed best, and due to space constraints, we count the number of classifier-metric combinations where the ablation performed worse than the full method, and calculate the counterfactual ranks, that is we calculate the rankings with the cWGAN results replaced by the ablation results. There are a total of 15 classifier-metric combinations. For comparison, our method achieved mean ranks of 3.1 and 2.4 on the Home Equity and Kaggle datasets, respectively, and was the best performing method in terms of mean rank on both, meaning it achieved a rank of mean ranks of 1.

\begin{table}[!htb]%
\caption[Ablation study results]{Ablation study results.}
\label{tab:ablation}
\begin{adjustbox}{max width =1.0 \textwidth}%
{\small
\begin{onehalfspacing}
\begin{tabular}{lllllllll}
\toprule
Ablated element & \multicolumn{2}{c}{WGANGP } & \multicolumn{2}{c}{AC} & \multicolumn{2}{c}{WGANGP \& AC} & \multicolumn{2}{c}{Treatment of cat. cols.} \\\cmidrule(lr){2-3} \cmidrule(lr){4-5} \cmidrule(lr){6-7} \cmidrule(lr){8-9}
Dataset & HomeEquity & Kaggle & HomeEquity & Kaggle &  HomeEquity & Kaggle &             HomeEquity & Kaggle \\
\midrule
\makecell[l]{Classifier-metric combinations\\ where ablation performs worse }&          11 &    7 &          13 &      4 &           11 &      9 &                      14 &    N/A \\
Counterfactual mean rank                           &         4.0 &    2.3 &         3.8 &    2.7 &          3.9 &    2.5 &                     3.5 &    N/A \\
Counterfactual rank of mean ranks                  &           3 &    1 &           3 &      2 &            3 &      2 &                       3 &    N/A \\
\bottomrule
\end{tabular}

\end{onehalfspacing}
}
\end{adjustbox}
\end{table}%

We find that all ablations perform worse in terms of mean rank and rank of mean ranks, except for not using the WGANGP loss on the Kaggle dataset. There, the counterfactual mean rank of 2.3 achieved with the Vanilla GAN loss was marginally better than the mean rank of 2.4 achieved with the WGANGP loss. Most ablations also perform worse on the majority of classifier-metric combinations, with the exceptions being not using the WGANGP loss, and not using an AC loss, which  outperformed our method on the Kaggle dataset on 8 and 11 of 15 classifier-metric combinations, respectively. However, when not using the AC loss, the mean rank was worse for the ablation, suggesting that the gains were generally minor and outweighed by the losses on the remaining four classifier-metric combinations. Furthermore, when ablating both the WGANGP and the AC losses together, the ablation performs worse on the majority of classifier-metric combinations, and achieves a worse mean rank. While developing our methodology on the UCI adult dataset, we often observed that without an AC loss, the Vanilla GAN loss could lead to mode collapse. This could explain why ablating either the WGANGP loss or the AC loss individually can improve performance slightly, while ablating both does not.

We conclude that each of the major elements of our method is useful and important for the performance as the ablations generally performed worse. However, the ablations also suggest that there is still potential in fine-tuning our method. In particular, which GAN loss to use, whether to use the AC loss, and the AC loss hyperparameters might be important hyperparameters to tune.

\FloatBarrier
\section{Conclusion}
We proposed cWGAN-based oversampling, and compared it against five benchmark oversampling techniques as well as the baseline of not oversampling. We performed the evaluation on seven real-world credit scoring datasets with five different classification algorithms using three metrics of classification performance. We find that cWGAN-based oversampling compares favourably to other oversampling methods. Our  method outperforms Random Oversampling on five out of seven datasets. It also outperforms SMOTE and the three tested SMOTE variants on four out of seven datasets, and on a fifth dataset, our method ties with SMOTE and outperforms the SMOTE variants. 

However, not oversampling performs better than any of the considered oversampling methods on five out of seven datasets. This indicates that in the context of credit scoring, class imbalance might not always be worth addressing. This could be due to the fact that many credit scoring datasets are mostly linear, which might make oversampling unnecessary. 

We find that on the only two strongly non-linear datasets in our evaluation, oversampling does improve performance and cWGAN is the best performing method. Furthermore, when using a high-performance, non-linear classification algorithm, such as Random Forest or Gradient Boosting, cWGAN performed well. Thus, we conclude that our proposed method is a powerful addition to the oversampling toolbox and might work especially well for strongly non-linear datasets.

Future research could repeat this benchmark on more datasets that are strongly non-linear to confirm our findings that in those settings, oversampling yields the largest benefits and that cWGAN is the best performing oversampling method. Further interesting research directions would be to consider datasets that are more heavily imbalanced, either by artificially undersampling the minority class or by considering new datasets, and to evaluate the effects of relative and absolute class imbalance. Additionally, our cWGAN-based oversampling method could be developed further to identify better default hyperparameter settings and key hyperparameters to tune, which could improve performance considerably. Finally, using cWGAN-based oversampling could be tested in other domains that face imbalanced classification problems with tabular data.

\section{Acknowledgments}
We would like to thank Johannes Haupt for valuable comments on GAN architectures and their implementations, and providing valuable demo codes in this GitHub repository at \url{https://github.com/johaupt/GANbalanced}.

\FloatBarrier
\newpage
\setlength{\parskip}{.3em}
{\singlespacing \small
\bibliography{cgan_oversampling}
}
\newpage

\appendix
\section{Oversampling benchmark - full results}%
\subsection{Ranks}%
\begin{table}[!htbp]%
\begin{adjustbox}{width = 0.90 \textwidth, max height = 0.90 \textheight}%
\centering%
\rotatebox{90}{
\begin{minipage}{\textheight}%
\caption[Appendix: Ranking of oversampling performance for each oversampling method per dataset for all classifiers]{Ranking of oversampling performance for each oversampling method per dataset for all classifiers. Best ranks reported in bold. Sorted by best overall rank per dataset.}
\resizebox{\textwidth}{!}{
\label{tab:oversampling_ranks}
\begin{tabular}{llllllllllllllllll}
\toprule
           & Classifier & \multicolumn{3}{l}{RandomForest} & \multicolumn{3}{l}{Logit \kern 0.8em} & \multicolumn{3}{l}{GradientBoosting} & \multicolumn{3}{l}{KNN \kern 1em} & \multicolumn{3}{l}{DecisionTree} &       Overall \\
           & Metric &      AUC-ROC &      AUC-PR & Brier\kern 1.2em &           AUC-ROC &      AUC-PR & Brier\kern 1.2em &          AUC-ROC &      AUC-PR & Brier\kern 1.2em &       AUC-ROC &      AUC-PR & Brier\kern 1.2em &      AUC-ROC &      AUC-PR & Brier\kern 1.2em &     Mean Rank \\
Dataset & Method &              &             &                  &                   &             &                  &                  &             &                  &               &             &                  &              &             &                  &               \\
\midrule
Germany & None &            2 &           2 &       \textbf{1} &        \textbf{1} &  \textbf{1} &       \textbf{1} &       \textbf{1} &  \textbf{1} &       \textbf{1} &             5 &  \textbf{1} &       \textbf{1} &            5 &           6 &                4 &  \textbf{2.2} \\
           & SMOTE &   \textbf{1} &  \textbf{1} &                2 &                 3 &           2 &                4 &                5 &           6 &                4 &             6 &           5 &                6 &            2 &           2 &                3 &           3.5 \\
           & B-SMOTE &            3 &           3 &                4 &                 2 &           5 &                6 &                6 &           7 &                5 &    \textbf{1} &           3 &                5 &   \textbf{1} &  \textbf{1} &                2 &           3.6 \\
           & Random &            5 &           5 &                5 &                 4 &           3 &                6 &                3 &           2 &                6 &             7 &           7 &                4 &            3 &           3 &       \textbf{1} &           4.3 \\
           & ADASYN &            4 &           4 &                3 &                 5 &           4 &                7 &                4 &           5 &                3 &             2 &           2 &                7 &            6 &           5 &                7 &           4.5 \\
           & cWGAN &            7 &           7 &                6 &                 6 &           7 &                2 &                2 &           4 &                2 &             4 &           4 &                2 &            7 &           7 &                5 &           4.8 \\
           & SMOTENC &            6 &           6 &                7 &                 7 &           6 &                3 &                7 &           3 &                7 &             3 &           6 &                3 &            4 &           4 &                6 &           5.2 \\
\midrule
HomeEquity & cWGAN &   \textbf{1} &  \textbf{1} &       \textbf{1} &                 7 &           7 &                2 &                3 &           2 &                2 &             5 &           6 &                6 &            2 &  \textbf{1} &       \textbf{1} &  \textbf{3.1} \\
           & SMOTE &            3 &           3 &                3 &                 3 &           3 &                5 &                6 &           4 &                3 &             4 &           2 &                2 &   \textbf{1} &           2 &                3 &  \textbf{3.1} \\
           & None &            4 &           2 &                7 &        \textbf{1} &  \textbf{1} &       \textbf{1} &       \textbf{1} &  \textbf{1} &       \textbf{1} &             7 &           7 &                7 &            6 &           3 &                2 &           3.4 \\
           & B-SMOTE &            5 &           4 &                4 &                 4 &           4 &                6 &                4 &           5 &                6 &             3 &           3 &                4 &            3 &           4 &                5 &           4.3 \\
           & ADASYN &            6 &           6 &                5 &                 5 &           2 &                7 &                5 &           6 &                4 &             2 &  \textbf{1} &       \textbf{1} &            4 &           7 &                7 &           4.5 \\
           & SMOTENC &            2 &           4 &                2 &                 6 &           6 &                3 &                7 &           7 &                7 &    \textbf{1} &           4 &                5 &            5 &           6 &                6 &           4.7 \\
           & Random &            7 &           7 &                6 &                 2 &           5 &                4 &                2 &           3 &                5 &             6 &           5 &                3 &            7 &           5 &                4 &           4.7 \\
\midrule
Kaggle & cWGAN &            2 &           2 &       \textbf{1} &                 5 &           4 &                2 &                3 &           3 &                2 &    \textbf{1} &  \textbf{1} &                2 &            4 &  \textbf{1} &                3 &  \textbf{2.4} \\
           & None &   \textbf{1} &  \textbf{1} &       \textbf{1} &                 6 &           6 &       \textbf{1} &                2 &  \textbf{1} &       \textbf{1} &             5 &           2 &       \textbf{1} &            5 &           2 &                2 &           2.5 \\
           & B-SMOTE &            3 &           4 &                3 &                 2 &           2 &                3 &                4 &           5 &                3 &             2 &           3 &                3 &            2 &           3 &                4 &           3.1 \\
           & Random &            4 &           3 &                2 &        \textbf{1} &  \textbf{1} &                4 &       \textbf{1} &           2 &                6 &             6 &           6 &                5 &            6 &           5 &       \textbf{1} &           3.5 \\
           & SMOTE &            5 &           5 &                4 &                 3 &           3 &                5 &                5 &           4 &                4 &             2 &           4 &                4 &   \textbf{1} &           3 &                5 &           3.8 \\
           & ADASYN &            6 &           6 &                5 &                 4 &           5 &                6 &                6 &           6 &                5 &             4 &           5 &                6 &            3 &           6 &                6 &           5.3 \\
\midrule
P2P & None &            3 &           3 &       \textbf{1} &                 2 &  \textbf{1} &       \textbf{1} &                2 &           2 &       \textbf{1} &             6 &           4 &       \textbf{1} &            4 &           4 &                3 &  \textbf{2.5} \\
           & cWGAN &            2 &           2 &       \textbf{1} &                 3 &           3 &       \textbf{1} &                3 &           3 &                2 &             5 &           3 &                2 &            7 &           7 &                2 &           3.1 \\
           & B-SMOTE &            4 &           6 &                3 &                 4 &           4 &                3 &                4 &           4 &                3 &    \textbf{1} &  \textbf{1} &                5 &   \textbf{1} &  \textbf{1} &                4 &           3.2 \\
           & Random &   \textbf{1} &  \textbf{1} &                2 &        \textbf{1} &           2 &                5 &       \textbf{1} &  \textbf{1} &                7 &             7 &           7 &                3 &            6 &           5 &       \textbf{1} &           3.3 \\
           & SMOTE &            4 &           4 &                4 &                 5 &           5 &                4 &                5 &           5 &                5 &             3 &           5 &                6 &            3 &           3 &                5 &           4.4 \\
           & SMOTENC &            7 &           7 &                6 &                 7 &           7 &                2 &                7 &           7 &                6 &             2 &           2 &                4 &            2 &           2 &                7 &           5.0 \\
           & ADASYN &            6 &           5 &                5 &                 6 &           6 &                6 &                6 &           6 &                4 &             4 &           6 &                7 &            5 &           6 &                6 &           5.6 \\
\midrule
PAKDD & None &            3 &  \textbf{1} &                2 &        \textbf{1} &  \textbf{1} &       \textbf{1} &       \textbf{1} &           2 &       \textbf{1} &    \textbf{1} &  \textbf{1} &       \textbf{1} &            7 &           7 &                2 &  \textbf{2.1} \\
           & cWGAN &            2 &           2 &       \textbf{1} &                 4 &           5 &                2 &                3 &           3 &                2 &             2 &           2 &                2 &            5 &           5 &       \textbf{1} &           2.7 \\
           & Random &   \textbf{1} &           3 &                6 &                 2 &           2 &                4 &                2 &  \textbf{1} &                7 &             4 &           4 &                4 &            2 &           2 &                3 &           3.1 \\
           & B-SMOTE &            4 &           4 &                3 &                 3 &           5 &                5 &                4 &           4 &                3 &             5 &           6 &                6 &            4 &           4 &                5 &           4.3 \\
           & SMOTE &            4 &           5 &                4 &                 6 &           3 &                6 &                5 &           5 &                4 &             6 &           5 &                5 &            3 &           3 &                4 &           4.5 \\
           & SMOTENC &            7 &           7 &                7 &                 7 &           7 &                3 &                7 &           7 &                6 &             3 &           3 &                3 &   \textbf{1} &  \textbf{1} &                7 &           5.1 \\
           & ADASYN &            6 &           6 &                5 &                 4 &           3 &                7 &                6 &           6 &                5 &             7 &           7 &                7 &            6 &           6 &                6 &           5.8 \\
\midrule
Taiwan & None &            2 &  \textbf{1} &       \textbf{1} &                 2 &  \textbf{1} &       \textbf{1} &                3 &           2 &       \textbf{1} &    \textbf{1} &  \textbf{1} &       \textbf{1} &            3 &           2 &                2 &  \textbf{1.6} \\
           & cWGAN &            3 &           2 &                2 &                 6 &           3 &                2 &                2 &           3 &                2 &             2 &           2 &                2 &   \textbf{1} &  \textbf{1} &                3 &           2.4 \\
           & Random &   \textbf{1} &           3 &                3 &        \textbf{1} &           2 &                3 &       \textbf{1} &  \textbf{1} &                7 &             4 &           4 &                6 &            4 &           3 &       \textbf{1} &           2.9 \\
           & SMOTE &            4 &           4 &                4 &                 4 &           4 &                4 &                4 &           4 &                3 &             3 &           3 &                4 &            2 &           4 &                4 &           3.7 \\
           & ADASYN &            6 &           6 &                6 &                 3 &           5 &                6 &                6 &           5 &                5 &             7 &           6 &                7 &            6 &           6 &                6 &           5.7 \\
           & SMOTENC &            6 &           5 &                7 &                 7 &           6 &                5 &                7 &           6 &                4 &             5 &           5 &                3 &            7 &           7 &                7 &           5.8 \\
           & B-SMOTE &            5 &           7 &                5 &                 5 &           7 &                7 &                5 &           7 &                6 &             6 &           7 &                5 &            5 &           5 &                5 &           5.8 \\
\midrule
Thomas & None &            2 &           2 &       \textbf{1} &                 3 &           2 &       \textbf{1} &                4 &           5 &                2 &             2 &  \textbf{1} &       \textbf{1} &   \textbf{1} &  \textbf{1} &       \textbf{1} &  \textbf{1.9} \\
           & Random &   \textbf{1} &  \textbf{1} &                3 &                 2 &  \textbf{1} &                3 &       \textbf{1} &  \textbf{1} &                5 &             4 &           3 &                6 &            2 &           2 &                2 &           2.5 \\
           & cWGAN &            4 &           3 &                2 &                 6 &           6 &                2 &                2 &           3 &       \textbf{1} &             5 &           4 &                2 &            6 &           4 &                3 &           3.5 \\
           & B-SMOTE &            3 &           4 &                4 &                 5 &           4 &                5 &                5 &           2 &                4 &    \textbf{1} &           2 &                3 &            5 &           6 &                6 &           3.9 \\
           & SMOTE &            5 &           5 &                5 &        \textbf{1} &           3 &                4 &                3 &           4 &                3 &             6 &           7 &                5 &            4 &           5 &                4 &           4.3 \\
           & SMOTENC &            7 &           6 &                7 &                 7 &           7 &                7 &                6 &           6 &                6 &             7 &           5 &                4 &            3 &           3 &                5 &           5.7 \\
           & ADASYN &            6 &           7 &                6 &                 4 &           5 &                6 &                7 &           7 &                7 &             3 &           6 &                7 &            7 &           7 &                7 &           6.1 \\
\bottomrule
\end{tabular}

}%
\end{minipage}%
}%
\end{adjustbox}%
\end{table}%

\FloatBarrier
\subsection{Raw scores}

\begin{table}[!ht]%
\caption[Appendix: AUC-ROC for each oversampling method per dataset for all classifiers]{AUC-ROC oversampling results in numerical format for each oversampling method per dataset for all classifiers. Averaged over six runs. Standard deviations are reported in brackets. Best value per classifier and dataset in bold. Higher is better.}
\label{tab:oversampling_results_auc_roc}
\resizebox{\textwidth}{!}{%
{\small
\begin{tabular}{lllllll}
\toprule
           & Classifier &              RandomForest &         Logit \kern 0.8em &          GradientBoosting &             KNN \kern 1em &              DecisionTree \\
Dataset & Method &                           &                           &                           &                           &                           \\
\midrule
Germany & None &           0.7605 (0.0355) &  \textbf{0.7455} (0.0209) &  \textbf{0.7577} (0.0142) &           0.6556 (0.0371) &           0.6086 (0.0401) \\
           & Random &           0.7529 (0.0407) &           0.7371 (0.0317) &           0.7513 (0.0305) &           0.6236 (0.0504) &           0.6385 (0.0194) \\
           & SMOTE &  \textbf{0.7625} (0.0381) &           0.7420 (0.0225) &           0.7457 (0.0359) &           0.6508 (0.0171) &           0.6506 (0.0467) \\
           & SMOTENC &           0.7527 (0.0450) &           0.7271 (0.0308) &           0.7433 (0.0355) &           0.6603 (0.0523) &           0.6180 (0.0249) \\
           & ADASYN &           0.7586 (0.0292) &           0.7365 (0.0119) &           0.7509 (0.0222) &           0.6658 (0.0409) &           0.5983 (0.0465) \\
           & B-SMOTE &           0.7589 (0.0421) &           0.7421 (0.0215) &           0.7450 (0.0213) &  \textbf{0.6704} (0.0277) &  \textbf{0.6554} (0.0573) \\
           & cWGAN &           0.7494 (0.0419) &           0.7360 (0.0402) &           0.7520 (0.0330) &           0.6598 (0.0254) &           0.5889 (0.0563) \\
\midrule
HomeEquity & None &           0.9733 (0.0085) &  \textbf{0.7738} (0.0237) &  \textbf{0.9213} (0.0203) &           0.9114 (0.0193) &           0.7867 (0.0191) \\
           & Random &           0.9703 (0.0079) &           0.7725 (0.0229) &           0.9197 (0.0213) &           0.9175 (0.0190) &           0.7799 (0.0333) \\
           & SMOTE &           0.9738 (0.0108) &           0.7723 (0.0227) &           0.9048 (0.0285) &           0.9262 (0.0151) &  \textbf{0.8047} (0.0255) \\
           & SMOTENC &           0.9746 (0.0101) &           0.7657 (0.0249) &           0.9021 (0.0291) &  \textbf{0.9337} (0.0140) &           0.7908 (0.0311) \\
           & ADASYN &           0.9719 (0.0105) &           0.7695 (0.0246) &           0.9049 (0.0213) &           0.9307 (0.0185) &           0.7920 (0.0277) \\
           & B-SMOTE &           0.9730 (0.0125) &           0.7707 (0.0243) &           0.9059 (0.0224) &           0.9263 (0.0180) &           0.7921 (0.0318) \\
           & cWGAN &  \textbf{0.9761} (0.0093) &           0.7554 (0.0278) &           0.9119 (0.0204) &           0.9229 (0.0194) &           0.7935 (0.0234) \\
\midrule
Kaggle & None &  \textbf{0.8415} (0.0044) &           0.6981 (0.0112) &           0.8330 (0.0054) &           0.6702 (0.0125) &           0.6084 (0.0066) \\
           & Random &           0.8330 (0.0038) &  \textbf{0.7857} (0.0081) &  \textbf{0.8342} (0.0056) &           0.6653 (0.0122) &           0.5984 (0.0097) \\
           & SMOTE &           0.8299 (0.0029) &           0.7772 (0.0080) &           0.8204 (0.0034) &           0.6965 (0.0092) &  \textbf{0.6155} (0.0093) \\
           & ADASYN &           0.8283 (0.0038) &           0.7679 (0.0090) &           0.8175 (0.0030) &           0.6960 (0.0100) &           0.6092 (0.0093) \\
           & B-SMOTE &           0.8372 (0.0030) &           0.7805 (0.0092) &           0.8235 (0.0047) &           0.6965 (0.0121) &           0.6106 (0.0052) \\
           & cWGAN &           0.8412 (0.0048) &           0.7449 (0.0431) &           0.8304 (0.0059) &  \textbf{0.7054} (0.0165) &           0.6090 (0.0067) \\
\midrule
P2P & None &           0.7539 (0.0106) &           0.7630 (0.0125) &           0.7708 (0.0144) &           0.6028 (0.0090) &           0.5586 (0.0151) \\
           & Random &  \textbf{0.7654} (0.0140) &  \textbf{0.7635} (0.0133) &  \textbf{0.7743} (0.0148) &           0.5989 (0.0099) &           0.5564 (0.0074) \\
           & SMOTE &           0.7531 (0.0094) &           0.7553 (0.0119) &           0.7452 (0.0131) &           0.6102 (0.0177) &           0.5654 (0.0103) \\
           & SMOTENC &           0.7197 (0.0133) &           0.6849 (0.0086) &           0.7061 (0.0110) &           0.6213 (0.0103) &           0.5698 (0.0081) \\
           & ADASYN &           0.7528 (0.0100) &           0.7544 (0.0121) &           0.7428 (0.0133) &           0.6083 (0.0168) &           0.5582 (0.0111) \\
           & B-SMOTE &           0.7531 (0.0086) &           0.7584 (0.0111) &           0.7528 (0.0142) &  \textbf{0.6230} (0.0183) &  \textbf{0.5720} (0.0126) \\
           & cWGAN &           0.7542 (0.0129) &           0.7615 (0.0113) &           0.7689 (0.0131) &           0.6049 (0.0136) &           0.5537 (0.0140) \\
\midrule
PAKDD & None &           0.6203 (0.0094) &  \textbf{0.6181} (0.0061) &  \textbf{0.6316} (0.0079) &  \textbf{0.5641} (0.0107) &           0.5231 (0.0095) \\
           & Random &  \textbf{0.6210} (0.0074) &           0.6170 (0.0056) &           0.6307 (0.0066) &           0.5573 (0.0094) &           0.5327 (0.0108) \\
           & SMOTE &           0.6160 (0.0091) &           0.6122 (0.0066) &           0.6123 (0.0057) &           0.5497 (0.0135) &           0.5293 (0.0097) \\
           & SMOTENC &           0.5960 (0.0067) &           0.5715 (0.0038) &           0.5832 (0.0050) &           0.5578 (0.0088) &  \textbf{0.5346} (0.0064) \\
           & ADASYN &           0.6147 (0.0105) &           0.6128 (0.0066) &           0.6106 (0.0078) &           0.5495 (0.0078) &           0.5251 (0.0106) \\
           & B-SMOTE &           0.6160 (0.0089) &           0.6133 (0.0058) &           0.6142 (0.0077) &           0.5509 (0.0079) &           0.5282 (0.0073) \\
           & cWGAN &           0.6206 (0.0048) &           0.6128 (0.0100) &           0.6287 (0.0115) &           0.5631 (0.0104) &           0.5252 (0.0069) \\
\midrule
Taiwan & None &           0.7702 (0.0103) &           0.7685 (0.0149) &           0.7850 (0.0121) &  \textbf{0.7063} (0.0195) &           0.6132 (0.0088) \\
           & Random &  \textbf{0.7715} (0.0107) &  \textbf{0.7691} (0.0150) &  \textbf{0.7857} (0.0122) &           0.6859 (0.0166) &           0.6104 (0.0028) \\
           & SMOTE &           0.7611 (0.0091) &           0.7676 (0.0139) &           0.7722 (0.0162) &           0.6910 (0.0167) &           0.6133 (0.0090) \\
           & SMOTENC &           0.7566 (0.0103) &           0.7577 (0.0119) &           0.7644 (0.0160) &           0.6833 (0.0135) &           0.6071 (0.0065) \\
           & ADASYN &           0.7566 (0.0085) &           0.7677 (0.0135) &           0.7652 (0.0163) &           0.6804 (0.0127) &           0.6080 (0.0053) \\
           & B-SMOTE &           0.7575 (0.0065) &           0.7624 (0.0139) &           0.7672 (0.0175) &           0.6807 (0.0164) &           0.6083 (0.0086) \\
           & cWGAN &           0.7701 (0.0088) &           0.7620 (0.0285) &           0.7851 (0.0138) &           0.7052 (0.0201) &  \textbf{0.6162} (0.0053) \\
\midrule
Thomas & None &           0.6265 (0.0343) &           0.6528 (0.0646) &           0.6283 (0.0487) &           0.5800 (0.0601) &  \textbf{0.5709} (0.0346) \\
           & Random &  \textbf{0.6371} (0.0273) &           0.6530 (0.0691) &  \textbf{0.6439} (0.0505) &           0.5743 (0.0649) &           0.5441 (0.0475) \\
           & SMOTE &           0.6161 (0.0395) &  \textbf{0.6533} (0.0666) &           0.6285 (0.0512) &           0.5635 (0.0607) &           0.5402 (0.0480) \\
           & SMOTENC &           0.6082 (0.0434) &           0.5925 (0.0605) &           0.6208 (0.0420) &           0.5505 (0.0469) &           0.5410 (0.0518) \\
           & ADASYN &           0.6117 (0.0426) &           0.6477 (0.0773) &           0.6084 (0.0564) &           0.5791 (0.0581) &           0.5284 (0.0306) \\
           & B-SMOTE &           0.6233 (0.0440) &           0.6447 (0.0722) &           0.6258 (0.0392) &  \textbf{0.5915} (0.0600) &           0.5366 (0.0224) \\
           & cWGAN &           0.6197 (0.0370) &           0.5985 (0.0768) &           0.6348 (0.0420) &           0.5644 (0.0571) &           0.5350 (0.0570) \\
\bottomrule
\end{tabular}

}
}%
\end{table}%

\begin{table}[!htb]%
\caption[Appendix: AUC-PR for each oversampling method per dataset for all classifiers]{AUC-PR oversampling results in numerical format for each oversampling method per dataset for all classifiers. Averaged over six runs. Standard deviations are reported in brackets. Best value per classifier and dataset in bold. Higher is better.}
\label{tab:oversampling_results_auc_pr}
\resizebox{\linewidth}{!}{%
{\small
\begin{tabular}{lllllll}
\toprule
           & Classifier &              RandomForest &         Logit \kern 0.8em &          GradientBoosting &             KNN \kern 1em &              DecisionTree \\
Dataset & Method &                           &                           &                           &                           &                           \\
\midrule
Germany & None &           0.6144 (0.0552) &  \textbf{0.5943} (0.0541) &  \textbf{0.6037} (0.0197) &  \textbf{0.4571} (0.0559) &           0.3666 (0.0515) \\
           & Random &           0.5973 (0.0681) &           0.5845 (0.0641) &           0.5951 (0.0540) &           0.4186 (0.0673) &           0.3972 (0.0238) \\
           & SMOTE &  \textbf{0.6228} (0.0742) &           0.5856 (0.0556) &           0.5786 (0.0666) &           0.4337 (0.0384) &           0.4027 (0.0531) \\
           & SMOTENC &           0.5797 (0.0772) &           0.5712 (0.0686) &           0.5910 (0.0576) &           0.4317 (0.0572) &           0.3775 (0.0390) \\
           & ADASYN &           0.6038 (0.0550) &           0.5793 (0.0387) &           0.5802 (0.0438) &           0.4540 (0.0413) &           0.3668 (0.0424) \\
           & B-SMOTE &           0.6093 (0.0838) &           0.5761 (0.0551) &           0.5779 (0.0369) &           0.4517 (0.0550) &  \textbf{0.4099} (0.0394) \\
           & cWGAN &           0.5766 (0.0712) &           0.5606 (0.0845) &           0.5852 (0.0398) &           0.4454 (0.0583) &           0.3543 (0.0669) \\
\midrule
HomeEquity & None &           0.9089 (0.0260) &  \textbf{0.5242} (0.0538) &  \textbf{0.8129} (0.0355) &           0.8257 (0.0370) &           0.5215 (0.0358) \\
           & Random &           0.8928 (0.0252) &           0.5204 (0.0541) &           0.8004 (0.0423) &           0.8627 (0.0336) &           0.5082 (0.0525) \\
           & SMOTE &           0.9084 (0.0306) &           0.5228 (0.0503) &           0.7834 (0.0477) &           0.8759 (0.0297) &           0.5281 (0.0453) \\
           & SMOTENC &           0.9057 (0.0321) &           0.5134 (0.0542) &           0.7760 (0.0490) &           0.8676 (0.0334) &           0.5027 (0.0609) \\
           & ADASYN &           0.9003 (0.0324) &           0.5232 (0.0514) &           0.7817 (0.0405) &  \textbf{0.8841} (0.0331) &           0.4990 (0.0489) \\
           & B-SMOTE &           0.9057 (0.0361) &           0.5227 (0.0488) &           0.7831 (0.0422) &           0.8694 (0.0339) &           0.5148 (0.0658) \\
           & cWGAN &  \textbf{0.9278} (0.0228) &           0.5091 (0.0604) &           0.8009 (0.0443) &           0.8447 (0.0371) &  \textbf{0.5374} (0.0471) \\
\midrule
Kaggle & None &  \textbf{0.3659} (0.0142) &           0.2257 (0.0076) &  \textbf{0.3830} (0.0069) &           0.1811 (0.0095) &           0.1203 (0.0057) \\
           & Random &           0.3347 (0.0141) &  \textbf{0.3145} (0.0091) &           0.3781 (0.0072) &           0.1424 (0.0077) &           0.1163 (0.0074) \\
           & SMOTE &           0.3141 (0.0125) &           0.3013 (0.0069) &           0.3428 (0.0088) &           0.1708 (0.0068) &           0.1166 (0.0050) \\
           & ADASYN &           0.3059 (0.0138) &           0.2856 (0.0087) &           0.3386 (0.0084) &           0.1661 (0.0081) &           0.1117 (0.0052) \\
           & B-SMOTE &           0.3218 (0.0112) &           0.3063 (0.0089) &           0.3398 (0.0106) &           0.1778 (0.0100) &           0.1166 (0.0056) \\
           & cWGAN &           0.3644 (0.0118) &           0.2864 (0.0432) &           0.3690 (0.0116) &  \textbf{0.2139} (0.0188) &  \textbf{0.1206} (0.0037) \\
\midrule
P2P & None &           0.2952 (0.0180) &  \textbf{0.2959} (0.0216) &           0.3133 (0.0214) &           0.1507 (0.0037) &           0.1338 (0.0065) \\
           & Random &  \textbf{0.2988} (0.0200) &           0.2914 (0.0218) &  \textbf{0.3140} (0.0197) &           0.1457 (0.0049) &           0.1323 (0.0037) \\
           & SMOTE &           0.2657 (0.0131) &           0.2847 (0.0187) &           0.2721 (0.0116) &           0.1489 (0.0069) &           0.1348 (0.0054) \\
           & SMOTENC &           0.2259 (0.0117) &           0.2259 (0.0137) &           0.2366 (0.0145) &           0.1529 (0.0082) &           0.1350 (0.0033) \\
           & ADASYN &           0.2646 (0.0112) &           0.2832 (0.0178) &           0.2709 (0.0115) &           0.1485 (0.0078) &           0.1316 (0.0042) \\
           & B-SMOTE &           0.2634 (0.0112) &           0.2878 (0.0182) &           0.2823 (0.0125) &  \textbf{0.1547} (0.0083) &  \textbf{0.1384} (0.0070) \\
           & cWGAN &           0.2954 (0.0246) &           0.2913 (0.0213) &           0.3093 (0.0202) &           0.1515 (0.0044) &           0.1311 (0.0060) \\
\midrule
PAKDD & None &  \textbf{0.3487} (0.0132) &  \textbf{0.3338} (0.0125) &           0.3530 (0.0133) &  \textbf{0.2905} (0.0039) &           0.2654 (0.0052) \\
           & Random &           0.3461 (0.0119) &           0.3310 (0.0121) &  \textbf{0.3535} (0.0133) &           0.2855 (0.0033) &           0.2698 (0.0064) \\
           & SMOTE &           0.3421 (0.0134) &           0.3301 (0.0125) &           0.3315 (0.0073) &           0.2811 (0.0047) &           0.2679 (0.0079) \\
           & SMOTENC &           0.3213 (0.0082) &           0.3019 (0.0067) &           0.3074 (0.0044) &           0.2870 (0.0042) &  \textbf{0.2700} (0.0060) \\
           & ADASYN &           0.3401 (0.0148) &           0.3301 (0.0121) &           0.3300 (0.0080) &           0.2802 (0.0048) &           0.2657 (0.0055) \\
           & B-SMOTE &           0.3424 (0.0124) &           0.3295 (0.0119) &           0.3326 (0.0096) &           0.2806 (0.0031) &           0.2674 (0.0040) \\
           & cWGAN &           0.3482 (0.0089) &           0.3295 (0.0149) &           0.3502 (0.0170) &           0.2892 (0.0029) &           0.2658 (0.0042) \\
\midrule
Taiwan & None &  \textbf{0.5472} (0.0203) &  \textbf{0.5416} (0.0207) &           0.5616 (0.0160) &  \textbf{0.4245} (0.0147) &           0.2878 (0.0142) \\
           & Random &           0.5379 (0.0160) &           0.5397 (0.0199) &  \textbf{0.5619} (0.0168) &           0.3639 (0.0106) &           0.2874 (0.0097) \\
           & SMOTE &           0.5249 (0.0188) &           0.5377 (0.0196) &           0.5484 (0.0204) &           0.3782 (0.0164) &           0.2817 (0.0097) \\
           & SMOTENC &           0.5189 (0.0196) &           0.5293 (0.0232) &           0.5323 (0.0215) &           0.3591 (0.0151) &           0.2764 (0.0130) \\
           & ADASYN &           0.5163 (0.0173) &           0.5354 (0.0191) &           0.5356 (0.0186) &           0.3534 (0.0114) &           0.2775 (0.0070) \\
           & B-SMOTE &           0.4996 (0.0179) &           0.5217 (0.0192) &           0.5218 (0.0194) &           0.3414 (0.0142) &           0.2783 (0.0035) \\
           & cWGAN &           0.5448 (0.0162) &           0.5380 (0.0312) &           0.5577 (0.0200) &           0.4217 (0.0177) &  \textbf{0.2899} (0.0082) \\
\midrule
Thomas & None &           0.4408 (0.0760) &           0.4784 (0.0870) &           0.4553 (0.0661) &  \textbf{0.3713} (0.0658) &  \textbf{0.3286} (0.0471) \\
           & Random &  \textbf{0.4428} (0.0764) &  \textbf{0.4822} (0.0814) &  \textbf{0.4818} (0.0740) &           0.3466 (0.0735) &           0.3159 (0.0640) \\
           & SMOTE &           0.4224 (0.0908) &           0.4738 (0.0897) &           0.4606 (0.0822) &           0.3350 (0.0686) &           0.3090 (0.0559) \\
           & SMOTENC &           0.4103 (0.0816) &           0.4140 (0.0547) &           0.4484 (0.0774) &           0.3375 (0.0680) &           0.3100 (0.0585) \\
           & ADASYN &           0.4081 (0.0768) &           0.4652 (0.0901) &           0.4413 (0.0871) &           0.3362 (0.0634) &           0.3006 (0.0414) \\
           & B-SMOTE &           0.4287 (0.0840) &           0.4693 (0.0956) &           0.4749 (0.0718) &           0.3603 (0.0755) &           0.3042 (0.0366) \\
           & cWGAN &           0.4310 (0.0777) &           0.4166 (0.0769) &           0.4636 (0.0665) &           0.3428 (0.0578) &           0.3092 (0.0637) \\
\bottomrule
\end{tabular}

}
}%
\end{table}%

\begin{table}[!htb]%
\caption[Appendix: Brier score for each oversampling method per dataset for all classifiers]{Brier score oversampling results in numerical format for each oversampling method per dataset for all classifiers. Averaged over six runs. Standard deviations are reported in brackets. Best value per classifier and dataset in bold. Lower is better.}
\label{tab:oversampling_results_brier}
\resizebox{\linewidth}{!}{%
{\small
\begin{tabular}{lllllll}
\toprule
           & Classifier &              RandomForest &         Logit \kern 0.8em &          GradientBoosting &             KNN \kern 1em &              DecisionTree \\
Dataset & Method &                           &                           &                           &                           &                           \\
\midrule
Germany & None &  \textbf{0.1720} (0.0134) &  \textbf{0.1811} (0.0088) &  \textbf{0.1750} (0.0126) &  \textbf{0.2062} (0.0144) &           0.3438 (0.0256) \\
           & Random &           0.1766 (0.0137) &           0.2149 (0.0127) &           0.1930 (0.0119) &           0.2761 (0.0208) &  \textbf{0.3126} (0.0300) \\
           & SMOTE &           0.1734 (0.0117) &           0.2142 (0.0081) &           0.1860 (0.0140) &           0.2870 (0.0123) &           0.3164 (0.0480) \\
           & SMOTENC &           0.1829 (0.0140) &           0.2132 (0.0167) &           0.1935 (0.0160) &           0.2632 (0.0245) &           0.3577 (0.0375) \\
           & ADASYN &           0.1741 (0.0106) &           0.2161 (0.0064) &           0.1835 (0.0098) &           0.2871 (0.0205) &           0.3588 (0.0488) \\
           & B-SMOTE &           0.1747 (0.0146) &           0.2149 (0.0087) &           0.1868 (0.0111) &           0.2839 (0.0199) &           0.3155 (0.0490) \\
           & cWGAN &           0.1804 (0.0139) &           0.2007 (0.0156) &           0.1784 (0.0172) &           0.2249 (0.0122) &           0.3576 (0.0465) \\
\midrule
HomeEquity & None &           0.0564 (0.0043) &  \textbf{0.1282} (0.0021) &  \textbf{0.0770} (0.0088) &           0.0834 (0.0066) &           0.1241 (0.0083) \\
           & Random &           0.0561 (0.0043) &           0.1810 (0.0098) &           0.0961 (0.0064) &           0.0445 (0.0057) &           0.1285 (0.0093) \\
           & SMOTE &           0.0541 (0.0051) &           0.1813 (0.0087) &           0.0953 (0.0087) &           0.0417 (0.0065) &           0.1254 (0.0121) \\
           & SMOTENC &           0.0536 (0.0054) &           0.1808 (0.0091) &           0.0978 (0.0084) &           0.0542 (0.0094) &           0.1350 (0.0146) \\
           & ADASYN &           0.0555 (0.0050) &           0.1917 (0.0095) &           0.0960 (0.0074) &  \textbf{0.0385} (0.0067) &           0.1383 (0.0163) \\
           & B-SMOTE &           0.0547 (0.0060) &           0.1874 (0.0098) &           0.0975 (0.0068) &           0.0467 (0.0062) &           0.1300 (0.0193) \\
           & cWGAN &  \textbf{0.0528} (0.0046) &           0.1389 (0.0046) &           0.0790 (0.0069) &           0.0783 (0.0052) &  \textbf{0.1193} (0.0137) \\
\midrule
Kaggle & None &  \textbf{0.0525} (0.0012) &  \textbf{0.0599} (0.0016) &  \textbf{0.0514} (0.0014) &  \textbf{0.0625} (0.0018) &           0.1044 (0.0032) \\
           & Random &           0.0566 (0.0011) &           0.1813 (0.0008) &           0.1499 (0.0013) &           0.1212 (0.0030) &  \textbf{0.1003} (0.0019) \\
           & SMOTE &           0.0614 (0.0011) &           0.1867 (0.0009) &           0.0841 (0.0021) &           0.1175 (0.0024) &           0.1213 (0.0039) \\
           & ADASYN &           0.0623 (0.0011) &           0.2042 (0.0010) &           0.0886 (0.0022) &           0.1213 (0.0025) &           0.1242 (0.0036) \\
           & B-SMOTE &           0.0584 (0.0010) &           0.1633 (0.0009) &           0.0804 (0.0018) &           0.1046 (0.0023) &           0.1148 (0.0032) \\
           & cWGAN &  \textbf{0.0525} (0.0013) &           0.0907 (0.0103) &           0.0520 (0.0014) &           0.0645 (0.0016) &           0.1054 (0.0042) \\
\midrule
P2P & None &  \textbf{0.0898} (0.0014) &  \textbf{0.0895} (0.0020) &  \textbf{0.0882} (0.0018) &  \textbf{0.1097} (0.0017) &           0.1868 (0.0057) \\
           & Random &           0.0912 (0.0015) &           0.1997 (0.0014) &           0.1915 (0.0017) &           0.2330 (0.0043) &  \textbf{0.1826} (0.0027) \\
           & SMOTE &           0.1046 (0.0017) &           0.1978 (0.0016) &           0.1133 (0.0017) &           0.2961 (0.0072) &           0.2058 (0.0055) \\
           & SMOTENC &           0.1182 (0.0019) &           0.1804 (0.0016) &           0.1249 (0.0018) &           0.2364 (0.0058) &           0.2288 (0.0062) \\
           & ADASYN &           0.1052 (0.0015) &           0.1998 (0.0014) &           0.1128 (0.0016) &           0.3022 (0.0062) &           0.2103 (0.0071) \\
           & B-SMOTE &           0.1034 (0.0015) &           0.1902 (0.0015) &           0.1116 (0.0018) &           0.2711 (0.0071) &           0.2033 (0.0077) \\
           & cWGAN &  \textbf{0.0898} (0.0015) &  \textbf{0.0895} (0.0016) &           0.0886 (0.0015) &           0.1109 (0.0020) &           0.1852 (0.0064) \\
\midrule
PAKDD & None &           0.1844 (0.0028) &  \textbf{0.1848} (0.0014) &  \textbf{0.1823} (0.0024) &  \textbf{0.2156} (0.0057) &           0.3672 (0.0070) \\
           & Random &           0.1912 (0.0020) &           0.2396 (0.0013) &           0.2359 (0.0013) &           0.2993 (0.0045) &           0.3681 (0.0089) \\
           & SMOTE &           0.1882 (0.0028) &           0.2403 (0.0015) &           0.1936 (0.0013) &           0.3376 (0.0051) &           0.3696 (0.0063) \\
           & SMOTENC &           0.2082 (0.0031) &           0.2346 (0.0012) &           0.2216 (0.0014) &           0.2789 (0.0053) &           0.3918 (0.0052) \\
           & ADASYN &           0.1890 (0.0029) &           0.2483 (0.0018) &           0.1945 (0.0015) &           0.3519 (0.0034) &           0.3738 (0.0100) \\
           & B-SMOTE &           0.1881 (0.0030) &           0.2402 (0.0013) &           0.1933 (0.0015) &           0.3377 (0.0040) &           0.3714 (0.0060) \\
           & cWGAN &  \textbf{0.1843} (0.0025) &           0.1887 (0.0055) &           0.1828 (0.0021) &           0.2169 (0.0055) &  \textbf{0.3651} (0.0066) \\
\midrule
Taiwan & None &  \textbf{0.1356} (0.0050) &  \textbf{0.1353} (0.0059) &  \textbf{0.1325} (0.0053) &  \textbf{0.1592} (0.0078) &           0.2744 (0.0054) \\
           & Random &           0.1408 (0.0044) &           0.1837 (0.0039) &           0.1790 (0.0031) &           0.2429 (0.0081) &  \textbf{0.2683} (0.0049) \\
           & SMOTE &           0.1543 (0.0034) &           0.1839 (0.0039) &           0.1639 (0.0042) &           0.2306 (0.0092) &           0.3085 (0.0082) \\
           & SMOTENC &           0.1593 (0.0035) &           0.1905 (0.0029) &           0.1737 (0.0034) &           0.2282 (0.0050) &           0.3194 (0.0054) \\
           & ADASYN &           0.1591 (0.0033) &           0.1981 (0.0039) &           0.1738 (0.0038) &           0.2468 (0.0055) &           0.3158 (0.0101) \\
           & B-SMOTE &           0.1584 (0.0029) &           0.2036 (0.0033) &           0.1762 (0.0033) &           0.2399 (0.0072) &           0.3120 (0.0091) \\
           & cWGAN &           0.1358 (0.0043) &           0.1389 (0.0119) &           0.1327 (0.0055) &           0.1604 (0.0078) &           0.2758 (0.0064) \\
\midrule
Thomas & None &  \textbf{0.2008} (0.0123) &  \textbf{0.1906} (0.0161) &           0.1971 (0.0117) &  \textbf{0.2219} (0.0221) &  \textbf{0.3447} (0.0118) \\
           & Random &           0.2035 (0.0105) &           0.2282 (0.0146) &           0.2178 (0.0087) &           0.2988 (0.0292) &           0.3529 (0.0306) \\
           & SMOTE &           0.2120 (0.0086) &           0.2299 (0.0169) &           0.2111 (0.0125) &           0.2972 (0.0324) &           0.3811 (0.0334) \\
           & SMOTENC &           0.2162 (0.0071) &           0.2403 (0.0190) &           0.2183 (0.0089) &           0.2944 (0.0318) &           0.3881 (0.0351) \\
           & ADASYN &           0.2155 (0.0083) &           0.2376 (0.0183) &           0.2192 (0.0099) &           0.3046 (0.0280) &           0.4024 (0.0256) \\
           & B-SMOTE &           0.2108 (0.0092) &           0.2318 (0.0178) &           0.2129 (0.0086) &           0.2918 (0.0311) &           0.3953 (0.0385) \\
           & cWGAN &           0.2014 (0.0124) &           0.2273 (0.0191) &  \textbf{0.1909} (0.0123) &           0.2422 (0.0261) &           0.3691 (0.0202) \\
\bottomrule
\end{tabular}

}
}%
\end{table}%

\end{document}